%% file: main.tex
\documentclass[10pt,journal,compsoc]{IEEEtran}

\ifCLASSOPTIONcompsoc
  \usepackage[nocompress]{cite}
\else
  \usepackage{cite}
\fi

\ifCLASSINFOpdf
  
\else
  
\fi

\usepackage[table]{xcolor}
\usepackage{dirtree}
\usepackage{color}
\usepackage{colortbl}
\usepackage{amsmath}
\usepackage{array,graphicx,subfigure}
\usepackage{booktabs}
\usepackage{multirow}
\usepackage{amssymb}
\usepackage{bbding}
\usepackage{tikz}
\definecolor{grn}{rgb}{0,0.7,0}
\usepackage[colorlinks,linkcolor=grn,anchorcolor=grn,citecolor=grn,urlcolor=black,CJKbookmarks=True]{hyperref}
\usepackage{algorithm}
\usepackage{algpseudocode}

\newcommand{\keypoint}[1]{\vspace{0.1cm}\noindent\textbf{#1}}
\newcommand{\tabincell}[2]{\begin{tabular}{@{}#1@{}}#2\end{tabular}}

\newcommand{\cut}[1]{}

\newcommand{\etal}{\textit{et al}.~}
\newcommand{\ie}{\textit{i}.\textit{e}.}
\newcommand{\eg}{\textit{e}.\textit{g}.}
\newcommand{\etc}{\textit{etc}}

\newcommand{\blue}[1]{\textcolor{black}{#1}}
\newcommand{\magenta}[1]{\textcolor{black}{#1}}

\newcommand{\sota}{state-of-the-art~}

\newcommand{\md}{multimodal}

\newcommand{\MD}{Multimodal}
\newcommand{\MM}{Multimodal}

\newcommand{\vani}{\textit{Vanilla}}

\begin{document}

\title{Multimodal Learning with Transformers: \\ A Survey}

\author{
Peng~Xu,
Xiatian~Zhu,
and~David~A.~Clifton
\IEEEcompsocitemizethanks{\IEEEcompsocthanksitem This paper is accepted by IEEE TPAMI.
\IEEEcompsocthanksitem Peng Xu is with Tsinghua University.  Xiatian Zhu is with the University of Surrey. David A. Clifton is with the University of Oxford, UK, and also with Oxford Suzhou Centre for Advanced Research, Suzhou, PRC.
\IEEEcompsocthanksitem Corresponding author: David A. Clifton 
}
}


\IEEEtitleabstractindextext{%
\begin{abstract}
Transformer is a promising neural network learner, and has achieved great success in various machine learning tasks.
Thanks to the recent prevalence of multimodal applications and big data,
Transformer-based multimodal learning has become a hot topic in AI research.
This paper presents a comprehensive survey of Transformer techniques oriented at
multimodal data.
The main contents of this survey include:
(1) a background of multimodal learning, Transformer ecosystem, and the multimodal big data era,
{(2) a} {systematic} {review of \vani{} Transformer, Vision Transformer, and multimodal Transformers, from a geometrically topological perspective,}
(3) a review of multimodal Transformer applications, via
two important paradigms, \ie, for multimodal pretraining and for specific multimodal tasks,
(4) a summary of the common challenges and designs shared by the multimodal Transformer models and applications, and
(5) a discussion of open problems and potential research directions for the community.
\end{abstract}

}

\maketitle

\IEEEdisplaynontitleabstractindextext

\IEEEpeerreviewmaketitle

\input{tex-2/section-1-introduction}

\input{tex-2/section-2-background}

\input{tex-2/section-3-transformers}

\input{tex-2/section-4-applications}
\input{tex-2/section-5-challenges}

\input{tex-2/section-6-discussion}

\input{tex-2/section-7-conclusion}

\ifCLASSOPTIONcaptionsoff
  \newpage
\fi

\bibliographystyle{IEEEtran}
\bibliography{bib}
\vspace{-1.5cm}
 \begin{IEEEbiography}[{\includegraphics[height=0.9in,clip,keepaspectratio]{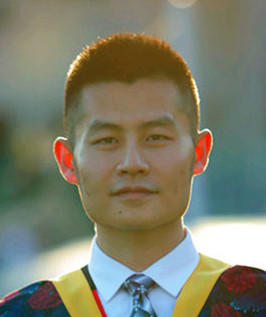}}]{Peng Xu} is a lecturer in the Department of Electronic Engineering, Tsinghua University.
 Previously, he was a postdoctoral research assistant in the Department of Engineering Science at the University of Oxford. 
 \end{IEEEbiography}
 \vspace{-2.2cm}
\begin{IEEEbiography}[{\includegraphics[height=0.9in,clip,keepaspectratio]{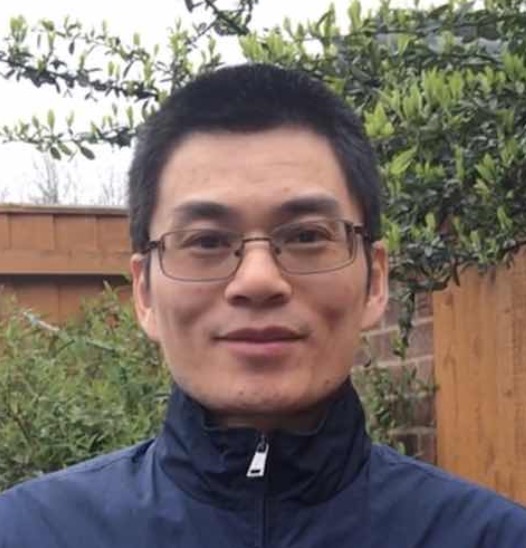}}]{Xiatian Zhu}
 is a Senior Lecturer at the Surrey Institute for People-Centred Artificial Intelligence, and Centre for Vision, Speech and Signal Processing (CVSSP), Faculty of Engineering and Physical Sciences, University of Surrey. 
 \end{IEEEbiography}
\vspace{-2.2cm}
\begin{IEEEbiography}[{\includegraphics[height=0.9in,clip,keepaspectratio]{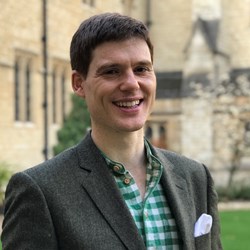}}]{David A. Clifton}
is a Professor of Clinical Machine Learning and leads the Computational Health Informatics (CHI) Lab in the Department of Engineering Science at the University of Oxford. 
 \end{IEEEbiography}

\clearpage

\appendix

\input{tex-2/appendix}

\end{document}

%% file: tex-2/section-1-introduction.tex
\IEEEraisesectionheading{\section{Introduction}\label{sec:introduction}}

The initial inspiration of Artificial Intelligence (AI)
is {to imitate human perception}, \eg, seeing, hearing, touching, smelling.
In general, a {modality} is often associated with a specific sensor that creates a unique communication channel, such as vision and language  \cite{baltruvsaitis2018multimodal}.
In humans,
a fundamental mechanism in our sensory perception is the ability to leverage multiple modalities of perception data collectively in order to engage ourselves properly with the world under dynamic unconstrained circumstances, with each modality serving as a distinct information source characterized by different statistical properties.
For example, an image gives the visual appearance of an ``elephants playing in water'' scene via thousands of pixels, whilst the corresponding text describes this moment with a sentence using discrete words.
%
Fundamentally, a multimodal AI system needs to ingest, interpret, and reason about multimodal information sources to realize similar human level perception abilities.
{Multimodal learning} (MML) is a general approach to building AI models that can extract and relate information from multimodal data \cite{baltruvsaitis2018multimodal}.

\begin{figure}[!t]
	\centering	
	\includegraphics[width=0.5\linewidth]{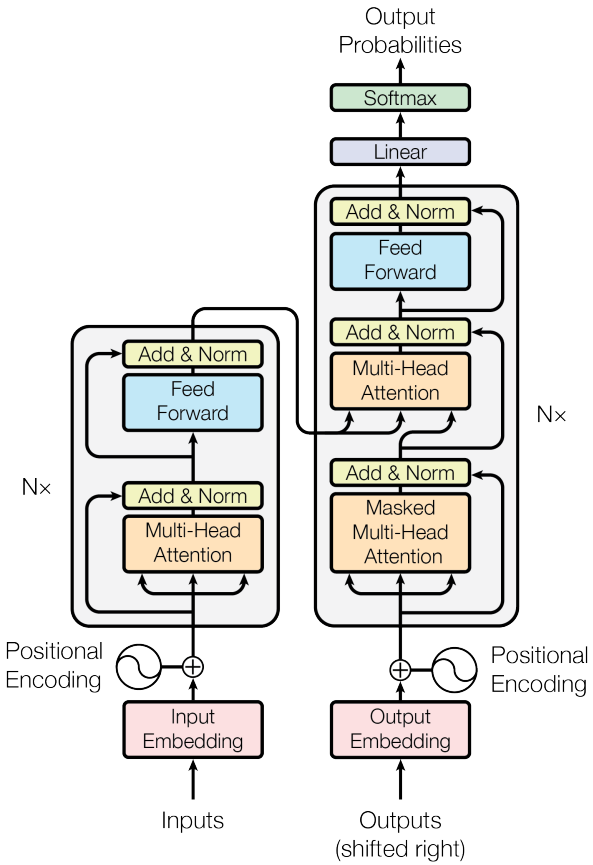}
	\caption{Overview of Transformer \cite{vaswani2017attention}.}
	\label{fig:transformer}
\end{figure}

This survey focuses on multimodal learning with Transformers \cite{vaswani2017attention} (as demonstrated in Figure \ref{fig:transformer}), inspired by their intrinsic advantages and scalability in modelling different modalities (\eg, language, visual, auditory) and tasks (\eg, language translation, image recognition, speech recognition) with fewer  modality-specific architectural assumptions (\eg, translation invariance and local grid attention bias in vision) \cite{jaegle2021perceiver}.
Concretely, the input to a Transformer could encompass one or multiple sequences of tokens, and each sequence's attribute (\eg, the modality label, the sequential order),
naturally allowing for MML without architectural modification \cite{devlin2018bert}.
Further, learning per-modal specificity and inter-modal correlation 
can be simply realized by controlling the input pattern of self-attention.
Critically, there is a recent surge of research attempts and activities across distinct disciplines  exploring the Transformer architectures, resulting in a large  number of novel MML methods being developed in recent years, along with significant and diverse advances in various areas
\cite{devlin2018bert,dosovitskiy2020image,carion2020end,sun2019videobert,chen2021speech}.
This calls for a timely review and summary of representative methods to enable researchers to understand the global picture of the MML field across related disciplines and more importantly to capture a holistic structured picture of  current achievements as well as major challenges.





\keypoint{Taxonomy}
For better readability and reachability from and across different disciplines, 
we adopt a {two-tier} structured taxonomy
based on the application and challenge dimensions respectively.
This has several benefits:
(1) Researchers with expertise in specific applications
can find 
those applications appropriate to their own research domain
before connecting to other related domains.
(2) Similar model designs and architectures developed in different domains
can be summarized in an abstract, formula-driven perspective so that the mathematical ideas of various models formed in different applications can be correlated and contrasted on common ground, crossing domain-specific restrictions.
Crucially, our taxonomy offers an interesting stereo-view of individual works
with the insights in both application specificity and formulation generality.
It is hoped that this can help to break down domain boundaries
and foster more effective idea communication and exchange across modalities.
By using the prompt modelling strategy  \cite{radford2021learning,li2022clip} as a basis for investigation, we also include 
the classical classification problem (\eg, image classification) -- usually regarded
as a single modality learning application in conventional MML surveys \cite{baltruvsaitis2018multimodal,zhang2020multimodal,rahate2022multimodal} --
as a special MML application.
This has the potential to significantly enrich MML, as the classification problem
is an AI topic amongst the most extensive studies in the literature \cite{hastie2009elements}.



\keypoint{Scope}
This survey will discuss the multimodality specific designs of Transformer architecture including, but not limited to, the following modalities:
RGB image \cite{dosovitskiy2020image}, depth image \cite{parida2022beyond}, \blue{multispectral image \cite{qingyun2021cross}}, video \cite{sun2019videobert}, audio/speech/music \cite{baevski2020wav2vec, nagrani2020speech2action, parida2022beyond},
table \cite{chen2020open}, scene graph/layout \cite{guo2021general,gupta2020layouttransformer,yang2021layouttransformer, Li_2022_CVPR_SGTR},
pose skeleton \cite{esser2021taming}, SQL \cite{cai2021sadga,song2022speech}, 
recipe \cite{salvador2021revamping}, programming language \cite{zhao2021proto},
sign language \cite{zhou2021improving, varol2021read, bull2021aligning},
point cloud \cite{zhao20213dvg},
symbolic knowledge (graph) \cite{marino2021krisp, ammanabrolu2021learning}, multimodal knowledge graph \cite{zhu2022multi}, sketch drawing \cite{xu2018sketchmate, xu2020deep,xu2020fine, vinker2022clipasso}, 3D object/scene \cite{fan2021faceformer, shin20193d, lin2021end}, document \cite{xu2020layoutlmv2,beltagy2020longformer}, programming code \cite{guo2020graphcodebert} and Abstract Syntax Tree (AST) -- a kind of graph \cite{zugner2021language}, optical flow \cite{gavrilyuk2020actor},
 medical knowledge (\eg, diagnosis code ontology \cite{shang2019pre}).
Note that this survey will not discuss the multimodal papers where Transformer is used simply  as the feature extractor without multimodal designs.


\keypoint{Related Surveys}
We relate this paper to existing surveys 
of the two specific dimensions MML and Transformers.
There exist a few MML surveys \cite{baltruvsaitis2018multimodal,zhang2020multimodal,rahate2022multimodal}.
In particular, \cite{baltruvsaitis2018multimodal} proposed a structured, acknowledged taxonomy by five challenges, which we also adopt
as part of our structure.
Unlike \cite{baltruvsaitis2018multimodal,zhang2020multimodal}, and \cite{rahate2022multimodal}, which review general machine learning models,
we instead focus on Transformer architectures and their self-attention mechanisms.
Several surveys dedicated to Transformers have been recently introduced, with a range of emphases including 
general Transformers \cite{lin2021survey},
efficient designs \cite{tay2020efficient},
visualization \cite{bracsoveanu2020visualizing},
computer vision tasks \cite{khan2021transformers,liu2021survey,han2020survey,xu2022transformers},
medical imaging \cite{shamshad2022transformers},
video tasks \cite{selva2022video}, and 
vision language pretraining \cite{ruan2021survey}.
While \cite{khan2021transformers,han2020survey,shamshad2022transformers,xu2022transformers} consider MML,
their reviews are somewhat  limited in the scope, taxonomy, and coverage.
%
To our knowledge, only a few surveys on video-language pretraining (VLP) \cite{ruan2021survey, chen2022vlp, li2022vision}  are relevant to MML.
However, VLP is only a subdomain of MML.
In this survey, we focus solely on the intersection of multimodal learning and Transformers.

\keypoint{{Features}}
To our knowledge, this paper is the first comprehensive review of the state of Transformer based multimodal machine learning.
The major features of this survey include 

{\textbf{(1)}} {We highlight that Transformers have the advantage that they can work in a modality-agnostic way. Thus, they are compatible with various modalities (and combinations of modalities). }
{To support this view, we, for the first time, offer an understanding of the intrinsic traits of Transformers in a multimodal context from a geometrically topological perspective.}
We suggest that self-attention be treated as a graph style modelling, which models the input sequence (both uni-modal and multimodal) as a fully-connected graph. 
Specifically, self-attention models the embedding of arbitrary tokens from an arbitrary modality as a graph node. 

{\textbf{(2)}} We discuss the key components of Transformers in a multimodal context as mathematically as possible.

{\textbf{(3)}} Based on Transformers, cross-modal interactions (\eg,
fusion, alignment)  are essentially  processed by self-attention
and its variants.
In this paper, we extract the mathematical essence and formulations of  Transformer based MML practices, from the perspective of self-attention designs.

\keypoint{{Contributions}}
Having presented our review of the landscape of multimodal learning, Transformer ecosystem, and multimodal big data era in Section \ref{sec:background},
we summarize our main contributions as the follows. 

{{\textbf{(1)}} In Section \ref{sec:multi-modal-transformer}, we present a 
{systematic} reviewing of \vani{} Transformer, Vision Transformer, and multimodal Transformers, from a geometrically topological perspective.} 

{\textbf{(2)}} We contribute a taxonomy for Transformer based MML from two complementary perspectives, \ie, application based and challenge based.
In Section \ref{sec:applications-and-representative-models},
we provide a review of multimodal Transformer applications, via
two important paradigms, \ie, for multimodal pretraining and for specific multimodal tasks.
In Section \ref{sec:challenges-and-designs}, 
we summarize the common challenges and designs shared by the various multimodal Transformer models and applications. 

{\textbf{(3)}} In Section \ref{sec:discussion-and-outlook}, we discuss current bottlenecks, existing  problems, and potential research directions for Transformer based MML.





%% file: tex-2/section-2-background.tex
\section{Background}
\label{sec:background}

\subsection{Multimodal Learning (MML)}
MML \cite{wu1999multimodal, baltruvsaitis2018multimodal, guo2019deep} has been an important research area in recent decades; an early multimodal application --
audio-visual speech recognition was studied in 1980s \cite{yuhas1989integration}.
MML is key to human societies.
The world we humans live in is a \md{} environment, thus both our observations and behaviours are  \md{} \cite{lazarus1976multimodal}. 
For instance, an AI navigation robot 
needs multimodal sensors to perceive the real-world environment \cite{feng2020deep, liu2021multimodal,moudgil2021soat}, \eg, camera, LiDAR, radar, ultrasonic, GNSS, HD Map, odometer.
Furthermore, human behaviours, emotions, events, actions, and humour are multimodal, thus various human-centred MML tasks are widely studied, including 
multimodal emotion recognition \cite{lv2021progressive}, multimodal event representation \cite{zellers2021merlot},
understanding multimodal humor \cite{hasan2021humor},
face-body-voice based video person-clustering \cite{brown2021face}, \etc.

Thanks to the development of the internet and a wide variety of intelligent devices in recent years, increasing amounts of multimodal data are being transmitted over the internet, thus an increasing number of multimodal application scenarios are emerging.
In modern life, we can see various multimodal applications, including commercial services (\eg, e-commerce/commodity retrieval \cite{yu2022commercemm}, vision-and-language navigation (VLN) \cite{chen2021topological, hong2021vln, zhang2021curriculum, qi2021road, chen2021semantic}), communication (\eg, lip reading \cite{ren2021learning}, sign language translation \cite{zhou2021improving, varol2021read}), human-computer interaction \cite{xu2022deep}, healthcare AI \cite{li2020behrt, li2020comparison}, surveillance AI \cite{xu2021deepchange}, \etc.

Moreover, in the era of Deep Learning, deep neural networks greatly promote the development of MML, and
Transformers \cite{vaswani2017attention} are a highly competitive architecture family,  bringing new challenges and opportunities to  MML.
\blue{In particular, the recent success of large language models and their multimodal derivatives \cite{tsimpoukelli2021multimodal, sung2022vl, alayrac2022flamingo, wang2022image, chen2022pali} further demonstrates the potential of Transformers in multimodal foundation models.}

\subsection{Transformers: a Brief History and Milestones}

Transformers are emerging as promising learners.
\vani{} Transformer \cite{vaswani2017attention} benefits from a self-attention mechanism, and is a breakthrough model for sequence-specific representation learning that was originally proposed for NLP, achieving the \sota{} on various NLP tasks.
{Following the great success of \vani{} Transformer, a lot of derivative models have been proposed, \eg, BERT \cite{devlin2018bert}, BART \cite{lewis2019bart}, GPT \cite{radford2018improving}, 
Longformer \cite{beltagy2020longformer},
Transformer-XL \cite{dai2019transformer},
XLNet \cite{yang2019xlnet}.}



Transformers currently  stand at the dominant position in NLP domains, and this motivates
researchers try to apply Transformers to other modalities, such as visual domains.
In early attempts for visual domain, the general pipeline 
is ``CNN features + standard Transformer encoder'', and researchers achieved BERT-style pretraining, via preprocessing raw images by resizing to a low resolution and reshaping into a 1D sequence \cite{chen2020generative}.

Vision Transformer (ViT) \cite{dosovitskiy2020image} is a seminal work that contributes an end-to-end solution by applying the encoder of Transformer to images. 
Both ViT and its variants have been widely  applied to various computer vision tasks, including
low-level tasks \cite{chen2021pre}, recognition \cite{touvron2021training}, detection \cite{beal2020toward}, segmentation \cite{liu2021swin}, \etc, and also work well for both supervised \cite{touvron2021training} and self-supervised \cite{chen2021empirical,caron2021emerging,bao2021beit} visual learning.
Moreover, some recently-released works provide further theoretical understanding for ViT, \eg, its internal representation robustness \cite{paul2021vision}, the continuous behaviour of its latent representation propagation \cite{raghu2021vision,cao2022understand}.

Motivated by the great success of Transformer, 
VideoBERT \cite{sun2019videobert} is a breakthrough work that is the first work to extend Transformer to the multimodal tasks. VideoBERT demonstrates the great potential of Transformer in multimodal context.
Following VideoBERT, a lot of Transformer based multimodal pretraining models (\eg, ViLBERT \cite{lu2019vilbert}, 
LXMERT \cite{tan2019lxmert}, VisualBERT \cite{li2019visualbert}, VL-BERT \cite{su2019vl}, UNITER \cite{chen2020uniter}, CBT \cite{sun2019learning}, Unicoder-VL \cite{li2020unicoder}, B2T2 \cite{alberti2019fusion}, VLP \cite{zhou2020unified}, 12-in-1 \cite{lu202012}, Oscar \cite{li2020oscar}, Pixel-BERT \cite{huang2020pixel}, ActBERT \cite{zhu2020actbert}, ImageBERT \cite{qi2020imagebert}, HERO \cite{li2020hero}, UniVL \cite{luo2020univl}) have become research topics of increasing interest  in the field of machine learning.

In 2021, CLIP \cite{radford2021learning} 
was proposed. It is a new milestone that uses \md{} pretraining to convert classification as a retrieval task that enables the pretrained models to tackle zero-shot recognition.
Thus, CLIP is a successful practice that makes full use of large-scale multimodal pretraining to enable zero-shot learning.
Recently, the idea of CLIP is further studied,
\eg, CLIP pretrained model based zero-shot semantic segmentation \cite{xu2021simple},
ALIGN \cite{jia2021scaling}, CLIP-TD \cite{wang2022clip}, ALBEF \cite{li2021alignbefore}, and CoCa \cite{yu2022coca}.




\subsection{\MM{} Big Data} 
{In the past decade, with the rapid development of internet applications such as social media and online retail,
massive multimodal 
datasets have been proposed, \eg,
Conceptual Captions \cite{sharma2018conceptual}, COCO \cite{lin2014microsoft}, VQA \cite{antol2015vqa}, Visual Genome \cite{krishna2017visual}, SBU Captions \cite{ordonez2011im2text}, Cooking312K \cite{sun2019videobert}, LAIT \cite{qi2020imagebert}, 
e-SNLI-VE \cite{kayser2021vil},
ARCH
\cite{gamper2021multiple}, Adversarial VQA \cite{li2021adversarial},
OTT-QA \cite{chen2020open},
MULTIMODALQA (MMQA) \cite{talmor2021multimodalqa},  
VALUE \cite{li2021value},
Fashion IQ \cite{wu2021fashion},
LRS2-BBC 
\cite{afouras2018deep},
ActivityNet \cite{krishna2017dense},
VisDial \cite{das2017visual}.}

Some emergent new trends among the recently released multimodal datasets are:   

{\textbf{(1)}} {Data scales are larger.}
Various recently released datasets are million-scale, \eg, 
Product1M \cite{zhan2021product1m},
Conceptual 12M \cite{changpinyo2021conceptual},
RUC-CAS-WenLan \cite{huo2021wenlan} (30M),
HowToVQA69M \cite{yang2021just},
HowTo100M \cite{miech2019howto100m},
ALT200M \cite{hu2021scaling},
LAION-400M \cite{schuhmann2021laion}.

{\textbf{(2)}} {More modalities.} 
In addition to the general modalities of vision, text, and audio, further diverse modalities are emerging, \eg, 
Pano-AVQA \cite{yun2021pano} -- the first large-scale spatial and audio-visual question
answering dataset on $360^{\circ}$ videos,
YouTube-360 (YT-360) \cite{morgado2020learning} ($360^{\circ}$ videos),
AIST++ \cite{li2021ai} (a new multimodal dataset of 3D
dance motion and music),
Artemis \cite{achlioptas2021artemis} (affective language for visual arts). In particular, MultiBench~\cite{liang2021multibench} provides a dataset including 10 modalities. 

{\textbf{(3)}}
{More scenarios.} 
In addition to common caption and QA datasets,
more applications and scenarios have been studied, \eg, 
CIRR \cite{liu2021image} (real-life images), 
Product1M \cite{zhan2021product1m},
Bed and Breakfast (BnB)  \cite{guhur2021airbert} (vision-and-language navigation),
M3A \cite{sawhney2021multimodal} (financial dataset),
X-World \cite{zhang2021x} (autonomous drive). 

{\textbf{(4)}}
{Tasks are more difficult.} 
Beyond the straightforward tasks,
more abstract multimodal tasks are proposed,
\eg, MultiMET \cite{zhang2021multimet} (a multimodal dataset for metaphor understanding),
Hateful Memes \cite{kiela2020hateful} (hate speech in multimodal memes). 

{\textbf{(5)}} {Instructional videos have become increasingly popular}, \eg, cooking video YouCookII \cite{zhou2018towards}.   Aligning a sequence of instructions to a video of someone carrying out a task is an example of a powerful pretraining pretext task \cite{malmaud2015s, sun2019videobert}.
\blue{Pretext tasks are pre-designed problems to force the models to learn representation by solving them.} 

\input{tex/datasets}
Similar to other deep neural network architectures, Transformers are also data hungry.
Therefore, their high-capacity models and multimodal big data basis co-create the prosperity of the Transformer based multimodal machine learning.
For instance, big data bring zero-shot learning capability to VLP Transformer models.

%% file: tex-2/section-3-transformers.tex
\section{{Transformers}}
\label{sec:multi-modal-transformer}

In this section, we use mathematical formulations to review the key techniques of \vani{} Transformer \cite{vaswani2017attention}, 
Vision Transformer \cite{dosovitskiy2020image}, and multimodal Transformers \footnote{In this survey, ``multimodal Transformer'' means ``Transformer in multimodal learning context''.}, including tokenized inputs, self-attention, multi-head attention, basic Transformer layers/blocks, \etc.
We highlight that \vani{} Transformers can be understood from a geometrically topological perspective \cite{bronstein2021geometric}, because due to the self-attention mechanism, given each tokenized input from any modalities, \vani{} self-attention (Transformer) can model it as a fully-connected graph in topological geometry space \cite{dwivedi2020generalization}.
Compared with other deep networks (for instance, CNN is restricted in the aligned grid spaces/matrices),
Transformers intrinsically have a more general and flexible modelling space.
This is a notable  advantage of Transformers for multimodal tasks.
Sections \ref{sec:vanilla-transformer}, \ref{sec:vit}, and \ref{sec:transformer-in-multimodal-context} will review the key designs of \vani{} Transformer, Vision Transformer, and multimodal Transformers, respectively.

\subsection{\vani{} Transformer}
\label{sec:vanilla-transformer}

\input{tex/transformer}

\vani{} Transformer has an encoder-decoder structure and is the origin of the  Transformer-based research field.
It takes tokenized input (see Section \ref{sec:tokenized-input}).
Both its encoder and decoder are stacked by the Transformer layers/blocks,
as demonstrated in Figure \ref{fig:transformer}.
Each block has two sub-layers, \ie, a multi-head self-attention (MHSA) layer (see Section \ref{sec:self-attention-and-multi-head-attention}) and a position-wise fully-connected feed-forward network (FFN) (see Section \ref{sec:ffn}).
To help the back propagation of the gradient, both MHSA and FFN use Residual Connection \cite{he2016deep} (given an input $x$, the residual connection of any mapping $f(\cdot)$ is defined as $x \gets f(x) + x$), followed by normalization layer.
Thus, assuming that the input tensor is $\mathbf{Z}$,  the output of MHSA and FFN sub-layers can be formulated as:
\begin{equation}
\label{eq:mhsa-and-ffn}
    \mathbf{Z} \gets N ( {sublayer} (\mathbf{Z}) + \mathbf{Z}),
\end{equation}
where ${sublayer}(\cdot)$ is the mapping  implemented by the sub-layer
itself and $N(\cdot)$ denotes normalization, \eg, $BN(\cdot)$ \cite{ioffe2015batch}, $LN(\cdot)$ \cite{ba2016layer}.

\keypoint{Discussion}
\magenta{There is an important unsolved problem that is post-normalization versus pre-normalization.} 
The original \vani{} Transformer uses post-normalization for each MHSA and FFN sub-layer.
However, if we consider this from the mathematical perspective, pre-normalization makes more sense {\cite{xiong2020layer}.} 
{This is similar to the basic principle of the theory of matrix, that normalization should be performed before projection, \eg, Gram–Schmidt process \footnote{\url{https://en.wikipedia.org/wiki/Gram\%E2\%80\%93Schmidt_process}}.}
This problem should be studied further by both theoretical research and experimental validation.

\subsubsection{Input Tokenization}
\label{sec:tokenized-input}

\keypoint{{Tokenization}}
\vani{} Transformer was originally proposed for machine translation as a sequence-to-sequence model, thus it is straightforward to take the vocabulary sequences as input.
As mentioned previously, the original self-attention can model an arbitrary input as a  fully-connected graph, independently of modalities.
Specifically, both \vani{} and variant Transformers take in the tokenized sequences, where each token can be regarded as a node of the graph.

\keypoint{{Special/Customized Tokens}}
{In Transformers,
various special/customized tokens can be semantically defined as place-holders in the token sequences,
\eg, mask token \texttt{[MASK]} \cite{devlin2018bert}.
Some common special tokens are summarized in appendix. Special tokens can be used in both uni-modal and multimodal Transformers.}

\keypoint{{Position Embedding}}
{  Position embeddings are added
to the token embeddings to retain positional information \cite{devlin2018bert}.
\vani{} Transformer uses sine and cosine functions to produce position embedding.
To date, various implementations of position embedding have been proposed.
The concrete solutions are outside  the focus of this survey.
}

\keypoint{Discussion}
{The main advantages of input tokenization include the following:} 

{\textbf{(1)}} Tokenization is a more general approach from a geometrically topological perspective, achieved by minimizing constraints caused by different modalities. 
{In general, every modality has intrinsic constraints on modelling.
For instance, sentences have  sequential structures that are well-suited by RNN, and photos are restricted in aligned grid matrices that CNN works well for.
Tokenization helps Transformers  inherently to process different modalities universally via irregular sparse structures.  Thus even \vani{} Transformer can encode multimodal inputs flexibly by just concatenation, weighted summation, even without any multimodal tailor-made modifications. }

{\textbf{(2)}} Tokenization is a more flexible approach to organize the input information via concatenation/stack, weighted summation, \etc.
\vani{} Transformer injects temporal information to the token embedding by summing position embedding.
For instance, when use Transformer to model free-hand sketch drawing \cite{xu2021multigraph}, each input token can integrate various drawing stroke patterns, \eg, stroke coordinates, stroke ordering, pen state (start/end). 

{\textbf{(3)}} {Tokenization is compatible with the task-specific customized tokens, \eg,  \texttt{[MASK]} token \cite{devlin2018bert} for Masked Language Modelling, \texttt{[CLASS]} token \cite{dosovitskiy2020image} for classification.} 



\keypoint{Discussion} 
\magenta{How to understand position embedding to Transformers is an open problem.}
It can be understood as a kind of implicit coordinate basis of feature space, to provide  temporal or spatial information to the Transformer.
For cloud point \cite{guo2021pct} and sketch drawing stroke \cite{xu2021multigraph}, their token element is already a coordinate, meaning that position embedding is optional, not necessary. 
Furthermore, position embedding can be regarded as a kind of general additional information.
In other words, from  a mathematical point of view, any additional information can be added, such as detail of the manner of position embedding, \eg,
the pen state of sketch drawing stroke \cite{xu2021multigraph},
cameras and viewpoints in surveillance \cite{he2021transreid}.
There is a comprehensive survey \cite{dufter2021position} discussing the position information in Transformers.
For both sentence structures (sequential) and general graph structures (sparse, arbitrary, and irregular),
position embeddings help Transformers to learn or encode the underlying structures.
Considered from the mathematical perspective of self-attention, \ie, scaled
dot-product attention,
attentions are invariant to the positions of words (in text) or nodes (in graphs), if  position embedding information is missing.
\blue{Thus, in most cases, position embedding is necessary for Transformers.}

\subsubsection{Self-Attention and Multi-Head Self-Attention}
\label{sec:self-attention-and-multi-head-attention}

The core component of \vani{} Transformer is the Self-Attention (SA) operation \cite{vaswani2017attention} that is also termed  ``Scaled Dot-Product Attention''.
Assume  that
$\mathbf{X} = [\mathbf{x}_1, \mathbf{x}_2, \cdots] \in \mathbb{R}^{N \times d}$ is an input sequence of $N$ elements/tokens,
and an optional preprocessing is positional encoding by point-wise summation $\mathbf{Z} \gets \mathbf{X} \oplus Position Embedding$ or concatenation $\mathbf{Z} \gets concat( \mathbf{X}, Position Embedding)$.

\keypoint{Self-Attention (SA)}
After preprocessing, embedding $\mathbf{Z}$ will
go through three projection matrices ($\mathbf{W}^{Q} \in \mathbb{R}^{d \times d_q}$, $\mathbf{W}^{K} \in \mathbb{R}^{d \times d_k}$, and $\mathbf{W}^{V} \in \mathbb{R}^{d \times d_v}$, $d_q = d_k$) to generate three embeddings $\mathbf{Q}$ (Query), $\mathbf{K}$ (Key), and $\mathbf{V}$ (Value):
\begin{equation}
\label{eq:projecting}
    \mathbf{Q} = \mathbf{Z} \mathbf{W}^{Q}, \mathbf{K} = \mathbf{Z} \mathbf{W}^{K}, \mathbf{V} = \mathbf{Z} \mathbf{W}^{V}.
\end{equation}
The output of self-attention is defined as
\begin{equation}
\label{eq:attention}
    \mathbf{Z}  = {SA} (\mathbf{Q}, \mathbf{K}, \mathbf{V}) = {Softmax}\left( \frac{\mathbf{Q} \mathbf{K}^\top}{\sqrt{d_q}} \right) \mathbf{V}.
\end{equation}
Given an input sequence, self-attention allows each element to attend to all the other elements, so that  self-attention encodes the input as a fully-connected graph. Therefore, the encoder of \vani{} Transformer can be regarded as a fully-connected GNN encoder, 
and the Transformer family has the non-local ability of global perception, similar to the Non-Local Network \cite{wang2018non}.

\keypoint{Masked Self-Attention (MSA)}
In practice, modification 
of  self-attention is needed to help the decoder of Transformer to learn  contextual dependence, to prevent positions from attending to subsequent positions, as
\begin{equation}
\label{eq:masked-attention}
    \mathbf{Z} = {MSA} (\mathbf{Q}, \mathbf{K}, \mathbf{V}) = {Softmax}\left( \frac{\mathbf{Q} \mathbf{K}^\top}{\sqrt{d_q}} \odot \mathbf{M} \right) \mathbf{V},
\end{equation}
where $\mathbf{M}$ is a masking matrix.
For instance, in GPT \cite{radford2018improving}, an upper triangular mask to enable look-ahead attention
where each token can only look at the past tokens.
Masking can be used in both encoder \cite{zhou2018end, xu2021multigraph} and decoder of Transformer, and has flexible implementations, \eg, 0-1 hard mask \cite{xu2021multigraph}, soft mask \cite{zhou2018end}.


In both uni-modal and multimodal practices,
specific masks are designed based on domain knowledge and prior knowledge.
Essentially, MSA is used to inject additional knowledge to Transformer models, \eg, \cite{xu2021multigraph,wang2019rat, cai2021sadga,wang2021sgeitl}. 

\keypoint{Multi-Head Self-Attention (MHSA)}
In practice, multiple self-attention sub-layers can be stacked in parallel and their concatenated outputs are fused by a projection matrix $\mathbf{W}$, to form a structure named Multi-Head Self-Attention:
\begin{equation}
\label{eq:multihead-attention}
    \mathbf{Z} = {MHSA} (\mathbf{Q}, \mathbf{K}, \mathbf{V}) = {concat} (\mathbf{Z}_{1}, \cdots, \mathbf{Z}_{H}) \textbf{W},
\end{equation}
where each head $\mathbf{Z}_{h} = {SA} (\mathbf{Q}_{h}, \mathbf{K}_{h} \mathbf{V}_{h})$ and $h \in [1, H]$,
and $\textbf{W}$ is a linear projection matrix.
The idea of MHSA is a kind of ensemble.
MHSA helps the model to jointly attend to information from multiple representation
sub-spaces.










\subsubsection{Feed-Forward Network}
\label{sec:ffn}
The output of the multi-head attention sub-layer will go through the position-wise Feed-Forward Network (FFN) that consists of successive linear layers with non-linear activation. 
For instance, a two-layer FFN can be formulated as
\begin{equation}
\label{eq:attention}
    FFN (\mathbf{Z}) =  \sigma (  \mathbf{Z} \mathbf{W}_{1} + \mathbf{b}_1) \mathbf{W}_{2} + \mathbf{b}_{2},
\end{equation}
where $\mathbf{W}_{1}$, $\mathbf{b}_{1}$, $\mathbf{W}_{2}$, and $\mathbf{b}_{2}$ denote the weights and biases of the two linear transformations, while $\sigma(\cdot)$ is non-linear activation, \eg, $\text{ReLU}(\cdot)$ \cite{glorot2011deep}, $GELU (\cdot)$ \cite{hendrycks2016gaussian}. In some Transformer literature, FFN is also termed  Multi-Layer Perceptron (MLP).

\subsection{Vision Transformer}
\label{sec:vit}


{Vision Transformer (ViT) \cite{dosovitskiy2020image}} 
{has an image-specific input pipeline in which the input image must be split into fixed-size (\eg, $16 \times 16$, $32 \times 32$) patches.}
After going through the linearly embedded layer and adding the position embeddings, all the patch-wise sequences will be encoded by a standard Transformer encoder.
Given an image $\mathbf{X} \in \mathbb{R}^{H \times W \times C}$ ($H$ height, $W$ width, $C$ channels), ViT needs to reshape $\mathbf{X}$ into a sequence of flattened 2D patches: $\mathbf{x}_{p} \in \mathbb{R}^\mathbf{N \times (P^{2} \cdot C)}$, where $(P \times P)$ is the patch resolution and $N = HW/P^2$.
To perform classification, a standard approach is to
prepend an extra learnable embedding ``classification token'' \texttt{[CLASS]} to the sequence of embedded patches:
\begin{equation}
\label{eq:attention}
    \textbf{Z} \gets concat(\texttt{[CLASS]}, \mathbf{X} \mathbf{W}),
\end{equation}
where $\mathbf{W}$ denotes the projection.



\input{tables/multi-modal-inputs}

\subsection{Multimodal Transformers}
\label{sec:transformer-in-multimodal-context}

Recently, a large number of  Transformers have been studied extensively for various multimodal tasks, and shown to be compatible with various modalities in both discriminative and generative tasks.

In this section, we will review the key techniques/designs of the existing multimodal Transformer models, from the perspectives of multimodal input (Section \ref{sec:multimodal-input}), self-attention variants (Section \ref{sec:self-attention-in-multimodal-context}), and network architectures (Section \ref{sec:architectures}).

\subsubsection{\MD{} Input}
\label{sec:multimodal-input}

The Transformer family is a general architecture that can be formulated as a type of general graph neural network.
Specifically, self-attention can process each input as a fully-connected graph, by attending to the global (non-local) patterns.
Therefore, this intrinsic trait helps Transformers can work in a  modality
agnostic pipeline that is compatible with various modalities by treating the embedding of each token as a node of the graph.

\keypoint{{Tokenization and Embedding Processing}}
Given an input from an arbitrary modality, 
users only need to perform two main steps, (1) tokenize the input, and (2) select an embedding space to represent the tokens, before inputting the data into Transformers.
In practice, both the tokenizing input and selecting embedding for the token are vital for Transformers but highly flexible, with many alternatives.
For instance, given an image, the solution of tokenizing and embedding is not unique.
Users can choose or design tokenization at multiple granularity levels -- coarse-grained vs. fine-grained.
\eg, use ROIs (obtained by an object detector) and CNN features as tokens and token embeddings \cite{lu2019vilbert}, use patches and linear projection as tokens and token embeddings \cite{dosovitskiy2020image}, or use graph node (obtained by object detector and graph generator) and GNN features as tokens and token embeddings \cite{yang2021multimodal}.
Given a tokenization plan, the subsequent embedding approaches can be diverse.
For example, for video input, a common  tokenization is to treat the non-overlapping
windows (down-sampled) over the video as tokens, and  their embeddings can then be extracted by various 3D CNNs, \eg, 
VideoBERT \cite{sun2019videobert}, CBT \cite{sun2019learning}, and UniVL \cite{luo2020univl} use S3D \cite{xie2017rethinking},
ActBERT uses ResNet-3D \cite{tran2018closer}.

Table~\ref{table:multi-modal-input} summarizes some common  practices of multimodal inputs for Transformers, including
 RGB,
 video,
 audio/speech/music,
 text,
 graph, \etc.

\keypoint{Discussion}
When considered from the perspective of geometric topology,
each of the modalities listed in Table~\ref{table:multi-modal-input} can be regarded as a graph. 
An RGB image is essentially a neat grid graph in the pixel space.
Both video and audio are clip/segment based graphs over a complex space involving  temporal and semantic patterns.
Both 2D and 3D drawing sketches \cite{xu2021multigraph, xu2022deep} are a kind of sparse graph if we consider their key points along the drawing strokes.
Similar to sketches, the human pose also is a kind of graph.
3D point cloud is a graph in which each coordinate is a node.
\magenta{Other abstract modalities also can be interpreted as graphs, \eg, 
source code \cite{guo2020graphcodebert},
data flow of source code \cite{guo2020graphcodebert},
table \cite{chen2020open},
SQL database schema \cite{song2022speech},
text question graph \cite{cai2021sadga}, and
electronic health records (EHRs) \cite{rasmy2021med}.}
\keypoint{Token Embedding Fusion}
In practice, 
Transformers allow each token position to contain multiple embeddings.
This is essentially a kind of early-fusion of embeddings, for both uni-modal and multimodal Transformer models. (This will be discussed further in subsequent sections.) 
The most common fusion is the token-wise summing of the multiple embeddings, \eg, a specific token embedding $\oplus$ position embedding.
Similar to the flexible tokenization, token embedding fusion is also flexible and widely applied to both uni-modal and multimodal Transformer applications.
In \cite{xu2021deepchange}, token-wise weighted summing is used to perform early-fusion of RGB and grey-scale images for multimodal surveillance AI.
In particular, token embedding fusion has an important role in multimodal Transformer applications as various embeddings can be fused by token-wise operators, \eg,
in VisualBERT \cite{li2019visualbert} and Unicoder-VL \cite{li2020unicoder}, segment embeddings are token-wise added to indicate which modality (vision or language) each token is from,
VL-BERT \cite{su2019vl} injects global visual context to linguistic domain by ``linguistic token embedding $\oplus$ full image visual feature
embedding'',
InterBERT \cite{lin2020interbert} adds location information for ROI by ``ROI embedding $\oplus$ location embedding'',
in ImageBERT \cite{qi2020imagebert}, five kinds of embeddings are fused ``image embedding $\oplus$ position embedding $\oplus$ linguistic embedding $\oplus$ segment embedding $\oplus$ sequence position embedding''.
















\input{tables/fusion-comparision}

\subsubsection{Self-Attention Variants in \MD{} Context}
\label{sec:self-attention-in-multimodal-context}

\begin{figure*}[!t]
	\centering
	\subfigure[]{
		\label{fig:fusion-1}
		\includegraphics[width=0.08\textwidth]{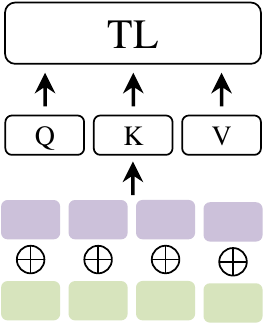}}
	\subfigure[]{
		\label{fig:fusion-2}
		\includegraphics[width=0.17\textwidth]{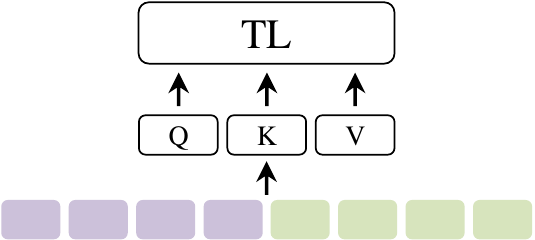}}
	\subfigure[]{
		\label{fig:fusion-3}
		\includegraphics[width=0.17\textwidth]{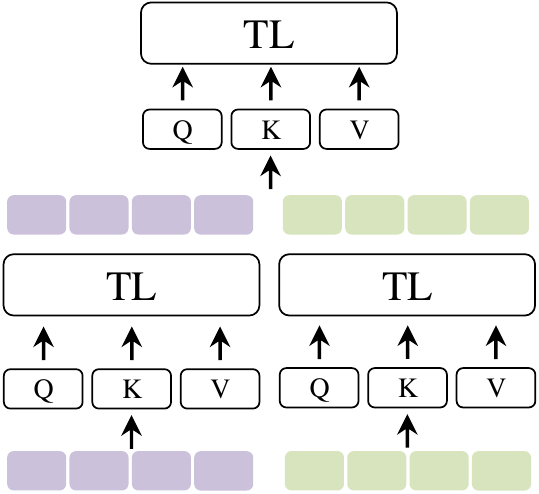}}
	\subfigure[]{
		\label{fig:fusion-4}
		\includegraphics[width=0.17\textwidth]{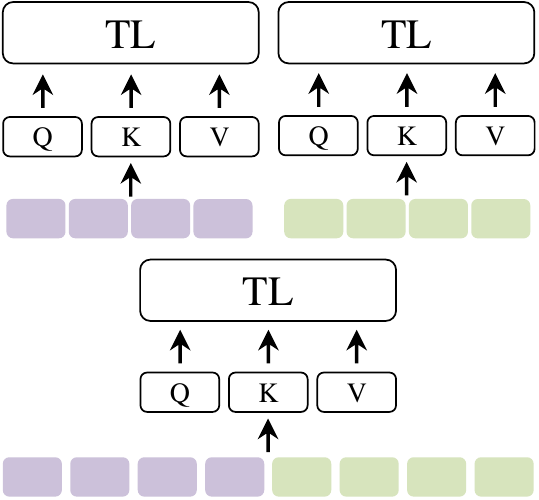}}
	\subfigure[]{
		\label{fig:fusion-5}
		\includegraphics[width=0.17\textwidth]{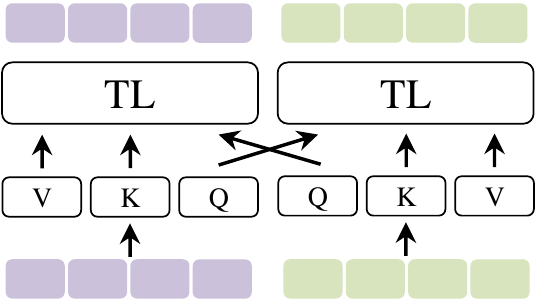}}
	\subfigure[]{
		\label{fig:fusion-6}
		\includegraphics[width=0.17\textwidth]{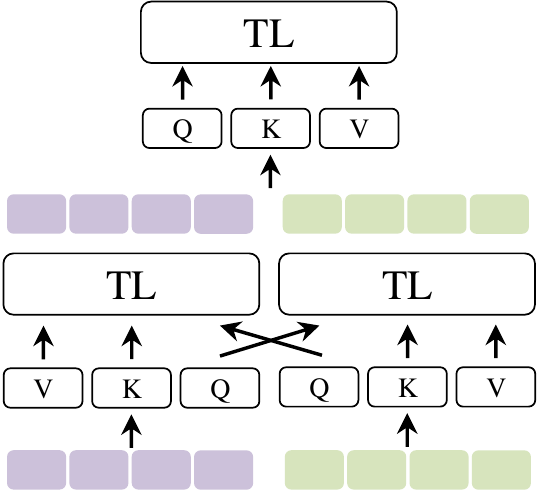}}
	\caption{Transformer-based cross-modal interactions\blue{: (a) Early Summation, (b) Early Concatenation, (c) Hierarchical Attention (multi-stream to one-stream), (d) Hierarchical Attention (one-stream to multi-stream), (e) Cross-Attention, and (f) Cross-Attention to Concatenation}.  ``Q'': Query embedding; ``K'': Key embedding; ``V'': Value embedding. ``TL'': Transformer Layer. Best viewed in colour. 
 }
	\label{fig:self-attention-fusion}
\end{figure*}

{In multimodal Transformers,
cross-modal interactions (\eg, fusion, alignment) are essentially processed by self-attention and its variants. 
Thus, in this section, we will review the main multimodal modelling practices of Transformers, from a perspective of self-attention designs, including
(1) early summation (token-wise, weighted),
(2) early concatenation,
(3) hierarchical attention (multi-stream to one-stream),
(4) hierarchical attention (one-stream to multi-stream),
(5) cross-attention, and
(6) cross-attention to concatenation.} See Table \ref{table:fusion-comparison} and Figure \ref{fig:self-attention-fusion}.


For brevity, we will state and compare the mathematical formulations in two-modality cases. Please note that all discussed self-attention and its variants are such flexible that can be extended to multiple modality cases.
{Specifically, the following formulations are modality-, tokenization-, and embedding- agnostic, as self-attention models the embedding of arbitrary token from arbitrary modality as a node of a graph.}   

Given inputs  $\mathbf{X}_\texttt{A}$ and $\mathbf{X}_\texttt{B}$ from two arbitrary modalities,
$\mathbf{Z}_{(\texttt{A})}$ and $\mathbf{Z}_{(\texttt{B})}$ denote their respective token embeddings.
Let $\mathbf{Z}$ denoting the token embedding (sequence) produced by the multimodal interactions.
$Tf(\cdot)$ stands for the processing of Transformer layers/blocks.


\keypoint{(1) Early Summation}
In practice, early summation \cite{gavrilyuk2020actor, xu2021deepchange} is a simple and effective multimodal interaction, where the token embeddings from multiple modalities can be weighted summed at each token position and then processed by Transformer layers:
\begin{equation}
\label{eq:early-summation}
   \mathbf{Z} \gets Tf(\alpha \mathbf{Z}_{(\texttt{A})} \oplus \beta \mathbf{Z}_{(\texttt{B})}) = {MHSA} (\mathbf{Q}_{(\texttt{AB})}, \mathbf{K}_{(\texttt{AB})}, \mathbf{V}_{(\texttt{AB})}), 
\end{equation}
where $\oplus$ is element-wise sum, and $\alpha$ and $\beta$ are weightings.
Concretely, $\mathbf{Q}_{(\texttt{AB})} = (\alpha \mathbf{Z}_{(\texttt{A})} \oplus \beta \mathbf{Z}_{(\texttt{B})}) \mathbf{W}^{Q}_{(\texttt{AB})}$, $\mathbf{K}_{(\texttt{AB})} = (\alpha \mathbf{Z}_{(\texttt{A})} \oplus \beta \mathbf{Z}_{(\texttt{B})}) \mathbf{W}^{K}_{(\texttt{AB})}$, and $\mathbf{V}_{(\texttt{AB})} = (\alpha \mathbf{Z}_{(\texttt{A})} \oplus \beta \mathbf{Z}_{(\texttt{B})}) \mathbf{W}^{V}_{(\texttt{AB})}$.
{Its main advantage is that it does not increase computational complexity.
However, its main disadvantage is due to the manually set weightings.} 
As discussed in Section \ref{sec:tokenized-input} and \ref{sec:multimodal-input}, summing position embedding is intrinsically a case of early summation.


\keypoint{(2) Early Concatenation}
Another straightforward solution is early concatenation \cite{sun2019videobert,   guo2020graphcodebert, shi2022learning, zheng2021fused} that the token embedding sequences from multiple modalities are concatenated and input into Transformer layers as
\begin{equation}
\label{eq:early-concatenation}
    \mathbf{Z} \gets Tf(\mathcal{C}(\mathbf{Z}_{(\texttt{A})}, \mathbf{Z}_{(\texttt{B})})).
\end{equation}
{Thus, all the multimodal token positions can be attended as a whole sequence, such that the positions of each modality can be encoded well by conditioning the context of other modalities.}
VideoBERT \cite{sun2019videobert} is the one of the first multimodal Transformer works, where video and text are fused via early concatenation {that can encode the global multimodal context well \cite{lin2020interbert}.
However, the longer sequence after concatenation will increase computational complexity.}
Early concatenation is also termed  ``all-attention'' or ``Co-Transformer'' \cite{zhan2021product1m}.


\keypoint{(3) Hierarchical Attention} (multi-stream to one-stream)
Transformer layers can be combined hierarchically to attend to the cross-modal interactions.
A common practice is that multimodal inputs are encoded by independent Transformer streams and their outputs are concatenated and fused by another Transformer \cite{li2021ai}:
\begin{equation}
\label{eq:hierarchical-attention-2-to-1}
    \mathbf{Z} \gets Tf_3(\mathcal{C}(Tf_1(\mathbf{Z}_{(\texttt{A})}), Tf_2(\mathbf{Z}_{(\texttt{B})}))).
\end{equation}
This kind of hierarchical attention is an implementation of late interaction/fusion, and  can be treated as a special case of early concatenation.


\keypoint{(4) Hierarchical Attention} (one-stream to multi-stream)
InterBERT \cite{lin2020interbert} is another good practice of hierarchical attention where concatenated multimodal inputs are encoded by a shared single-stream Transformer that is followed by two separate Transformer streams.
This flow can be formulated as
\begin{equation}
\label{eq:hierarchical-attention-1-to-2}
    \left\{ 
    \begin{aligned}
        \blue{\mathcal{C}(}\mathbf{Z}_{(\texttt{A})}, \mathbf{Z}_{(\texttt{B})}\blue{)} & \gets Tf_{1}(\mathcal{C}(\mathbf{Z}_{(\texttt{A})}, \mathbf{Z}_{(\texttt{B})})), \\
        \mathbf{Z}_{(\texttt{A})} & \gets Tf_{2}(\mathbf{Z}_{(\texttt{A})}), \\
        \mathbf{Z}_{(\texttt{B})} & \gets Tf_{3}(\mathbf{Z}_{(\texttt{B})}).
    \end{aligned}
    \right.
\end{equation}
{This method perceives the cross-modal interactions and meanwhile preserves the independence of uni-modal representation.}


\keypoint{(5) Cross-Attention}
For two-stream Transformers, if the $\mathbf{Q}$ (Query) embeddings are exchanged/swapped in a cross-stream manner, the cross-modal interactions can also be perceived.
This method is termed cross-attention or co-attention \cite{murahari2020large}, which was first proposed in VilBERT \cite{lu2019vilbert}:
\begin{equation}
\label{eq:cross-attention}
    \left\{ 
    \begin{aligned}
        \mathbf{Z}_{(\texttt{A})} & \gets {MHSA} (\mathbf{Q}_{\texttt{B}}, \mathbf{K}_{\texttt{A}}, \mathbf{V}_{\texttt{A}}), \\
        \mathbf{Z}_{(\texttt{B})} & \gets {MHSA} (\mathbf{Q}_{\texttt{A}}, \mathbf{K}_{\texttt{B}}, \mathbf{V}_{\texttt{B}}).
    \end{aligned}
    \right.
\end{equation}
{Cross-attention attends to each modality
conditioned on the other and does not cause higher computational complexity, however if considered for each modality, this method fails to perform  cross-modal attention globally and thus loses the whole context.
As discussed in \cite{lin2020interbert}, 
two-stream cross-attention can learn cross-modal interaction, whereas there is no self-attention to the self-context inside each modality.}


\keypoint{(6) Cross-Attention to Concatenation}
The two streams of cross-attention \cite{lu2019vilbert} can be further concatenated and processed by another Transformer to model the global context.
This kind of hierarchically cross-modal interaction is also widely studied
\cite{tsai2019multimodal,zhan2021product1m}, and {alleviates the drawback of cross-attention.}
\begin{equation}
\label{eq:cross-attention-to-concatenation}
    \left\{ 
    \begin{aligned}
        \mathbf{Z}_{(\texttt{A})} & \gets {MHSA} (\mathbf{Q}_{\texttt{B}}, \mathbf{K}_{\texttt{A}}, \mathbf{V}_{\texttt{A}}), \\
        \mathbf{Z}_{(\texttt{B})} & \gets {MHSA} (\mathbf{Q}_{\texttt{A}}, \mathbf{K}_{\texttt{B}}, \mathbf{V}_{\texttt{B}}), \\
        \mathbf{Z} & \gets Tf(\mathcal{C}(\mathbf{Z}_{(\texttt{A})}, \mathbf{Z}_{(\texttt{B})})).
    \end{aligned}
    \right.
\end{equation}

\keypoint{Discussion}
{All these aforementioned self-attention variants for multimodal interactions are modality-generic, and can be applied in flexible strategies and for multi-granular tasks.
Specifically, these interactions can be flexibly combined and nested.}
For instance, multiple cross-attention streams are used in hierarchical attention (one-stream to multi-stream) that in a two-stream decoupled model \cite{li2021scheduled} $Tf_2$ and $Tf_3$ of Eq. \ref{eq:hierarchical-attention-1-to-2} are implemented by cross-attention defined in Eq. \ref{eq:cross-attention}. 
{Moreover, they can be extended to multiple ($\geq 3$) modalities.}
TriBERT \cite{rahman2021tribert} is a tri-modal cross-attention (co-attention) for vision, pose, and audio, where given a Query embedding, its Key and Value embeddings are the concatenation from the other modalities. Cross-attention to concatenation is applied to three modalities (\ie, language,
video, and audio) in \cite{tsai2019multimodal}.


\subsubsection{Network Architectures}
\label{sec:architectures}

Essentially,
various multimodal Transformers work due to their internal multimodal attentions that are 
the aforementioned self-attention variants.
Meanwhile, as illustrated in Figure \ref{fig:self-attention-fusion}, these attentions determine the external network structures of the multimodal Transformers where they are embedded.  

In general, if we consider from the angle of network structures, (1) early summation and early concatenation work in single-stream, (2) cross-attention work in multi-streams, (3) hierarchical attention and  cross-attention to concatenation work in hybrid-streams.
Thus,
multimodal Transformers can be divided into
single-stream (\eg, Uniter \cite{chen2020uniter}, Visualbert \cite{li2019visualbert},  Vl-bert \cite{su2019vl} , Unified VLP \cite{zhou2020unified}), multi-stream (\eg, ViLBERT \cite{lu2019vilbert}, Lxmert \cite{tan2019lxmert},  ActBERT \cite{zhu2020actbert}),  hybrid-stream (\eg, InterBERT \cite{lin2020interbert}), \etc.

From the perspective of timing of interaction, these multimodal attentions fall into three categories, \ie, early interaction: early summation, early concatenation, and hierarchical attention (one-stream to multi-stream), late interaction: hierarchical attention (multi-stream to one-stream), or throughout interaction: cross-attention,  cross-attention to concatenation.


As demonstrated in Figure 2 in \cite{kim2021vilt}, the multimodal Transformer models have another architecture taxonomy based on the computational size of the components.
















%% file: tex/transformer.tex


%% file: tables/multi-modal-inputs.tex
\begin{table*}[!t]
\caption{ 
 Tokenization and token embedding comparison for multi-modal inputs for Transformers.
``ICD'': International Classification of Diseases. 
}
\label{table:multi-modal-input}
\begin{center}
\resizebox{\textwidth}{!}{
\begin{tabular}{ l |   l |  l | l}
\hline
Modalities & {Tokenization} & {Token Embeddings} & References \\
\hline
\hline
 RGB    &  {RoI}   &  {CNN embedding} &  ViLBERT \cite{lu2019vilbert}, LXMERT \cite{tan2019lxmert}, \\ 



 RGB    &  {patch}    & {linear projection}     & ViT \cite{dosovitskiy2020image} \\
\hline
\hline



 video & {clip of sampled frames} & {3D CNN embedding} & VideoBERT \cite{sun2019videobert}, CBT \cite{sun2019learning}, ActBERT \cite{zhu2020actbert} \\
 video & {sampled frame} & {2D CNN embedding}  & \cite{lin2021exploring} \\

 video & {voxel of sampled frames} & {linear projection} & VATT \cite{akbari2021vatt}  \\

 video & {patch of sampled frame} & {linear projection} &  MBT \cite{nagrani2021attention}  \\

 $360^{\circ}$ video  &  {clip of sampled frames}  & {3D CNN embedding} & AVSA \cite{morgado2020learning} \\
\hline 
\hline

 audio & {frame (mel-spectrogram)} & {CNN embedding}  & \cite{lin2021exploring}, AVSA \cite{morgado2020learning} \\

 audio &  {waveform segment}  &    {linear projection}   &  VATT \cite{akbari2021vatt} \\

 audio &  {spectrogram patch}   &    {linear projection}   &  MBT \cite{nagrani2021attention} \\




 speech/audio    &  {waveform segment}  & {1D-CNN (TCN) embedding}    & \cite{duquenne2021multimodal}, FaceFormer \cite{fan2021faceformer} \\

 speech &   {frame (mel-spectrogram)} & {linear projection and gated CNN embedding} & Meta-StyleSpeech \cite{min2021meta} \\

 speech & {frame (log-Mel filterbanks)} &  {linear projection} & AV-HuBERT \cite{shi2022learning}  \\

 speech/audio & {frame (log power spectrum)} & {linear projection} & VSET \cite{ramesh2021vset} \\

 speech & {frame (log-Mel filterbanks)} & {2D-CNN embedding} & FAT-MLM \cite{zheng2021fused}  \\



 music &  {frame (35-dim music feature)}   &  {linear projection}    &  FACT \cite{li2021ai} \\

\hline
\hline
 text    &  {word}  &   {learned embedding}   & \vani{} Transformer \cite{vaswani2017attention} \\

 text & {word} & {GNN embedding} & MGNNS \cite{yang2021multimodal} \\

\hline
\hline
 \magenta{SQL database schema}     &  {table node, column node}  &    {GNN embedding} & SpeechSQLNet \cite{song2022speech} \\

 \magenta{textual question-graph}     &  {word node}  &    {GNN embedding} & SADGA \cite{cai2021sadga} \\


 sketch    & {key point of stroke}   & {linear projection and learnable embedding} & Multi-Graph Transformer \cite{xu2021multigraph}    \\
 sketch    &  {patch of picture}  &  {linear projection}   & RVT \cite{mao2021towards} \\

 \magenta{table}     &  {cell}  &  {learned embedding} &  \cite{chen2020open} \\

 3D point cloud     &  {point}    &  {non-linear projection}  & Point Cloud Transformer
 \cite{guo2021pct} \\

 \magenta{source code}     &  {code}   & {learned embedding}    &  GraphCodeBERT \cite{guo2020graphcodebert} \\

 \magenta{data flow of source code}    &   {variable}   &   {learned embedding}   & GraphCodeBERT \cite{guo2020graphcodebert} \\

 pose    & {key point}   &  {GCN embedding}   &  TriBERT \cite{rahman2021tribert} \\

 electronic health records (EHRs)    &  {ICD code}  &  {GNN embedding}  & G-BERT \cite{shang2019pre} \\

 \magenta{electronic health records (EHRs)}    &  {ICD code}   &   {learned embedding}   & Med-BERT \cite{rasmy2021med} \\

 Gigapixel Whole Slide Images        &  {patch}   &    {CNN embedding}  & MCAT \cite{chen2021multimodal} \\

\hline
\end{tabular}
}
\end{center}

\end{table*}


%% file: tables/fusion-comparision.tex
\begin{table*}[!h]
\caption{ 
Self-attention variants for multi-modal interaction/fusion.
$\alpha$ and $\beta$ denote weightings.
``Att.'': Attention; ``Concat.''/``Con.'': Concatenation;
``Tfs'': Transformer layers.
$N_{(\texttt{A})}$ and $N_{(\texttt{B})}$ denote the token sequence lengths of two modalities.
}
\label{table:fusion-comparison}
\begin{center}
\resizebox{\textwidth}{!}{
\begin{tabular}{ l | c | c | c |  c | c }
\hline
Self-Attention & Definitions & Streams & Formulations & {Complexities}  & References  \\
\hline
\hline

Early Summation & token-wise sum before Tfs & 1 & $\mathbf{Z} \gets Tf(\alpha \mathbf{Z}_{(\texttt{A})} \oplus \beta \mathbf{Z}_{(\texttt{B})})$ & {$\mathcal{O}(N_{(\texttt{A})}^{2})$} &   \cite{gavrilyuk2020actor, xu2021deepchange}  \\

Early Concat. & token sequence concat. before Tfs & 1 & $\mathbf{Z} \gets Tf(\mathcal{C}(\mathbf{Z}_{(\texttt{A})}, \mathbf{Z}_{(\texttt{B})}))$ & {$\mathcal{O}((N_{(\texttt{A})} + N_{(\texttt{B})})^{2})$} &  \cite{sun2019videobert,   guo2020graphcodebert, shi2022learning, zheng2021fused} \\

Hierarchical Att. &  2-stream Tfs followed by concat. & 2 $\rightarrow$ 1& $\mathbf{Z} \gets Tf_3(\mathcal{C}(Tf_1(\mathbf{Z}_{(\texttt{A})}), Tf_2(\mathbf{Z}_{(\texttt{B})})))$ & {$\mathcal{O}((N_{(\texttt{A})} + N_{(\texttt{B})})^{2})$} &   \cite{li2021ai},  \\

Hierarchical Att. & early concat. followed by 2-stream Tfs & $1 \rightarrow$ 2& \blue{     $\left\{ 
    \begin{aligned}
        \blue{\mathcal{C}(}\mathbf{Z}_{(\texttt{A})}, \mathbf{Z}_{(\texttt{B})}\blue{)} & \gets Tf_{1}(\mathcal{C}(\mathbf{Z}_{(\texttt{A})}, \mathbf{Z}_{(\texttt{B})})), \\
        \mathbf{Z}_{(\texttt{A})} & \gets Tf_{2}(\mathbf{Z}_{(\texttt{A})}), \\
        \mathbf{Z}_{(\texttt{B})} & \gets Tf_{3}(\mathbf{Z}_{(\texttt{B})}).
    \end{aligned}
    \right.$}  & {$\mathcal{O}((N_{(\texttt{A})} + N_{(\texttt{B})})^{2})$}  &  \cite{lin2020interbert}  \\


Cross-Attention  & exchange query & 2 &  $\left\{ 
    \begin{aligned}
        \mathbf{Z}_{(\texttt{A})} & \gets {MHSA} (\mathbf{Q}_{\texttt{B}}, \mathbf{K}_{\texttt{A}}, \mathbf{V}_{\texttt{A}}) \\
        \mathbf{Z}_{(\texttt{B})} & \gets {MHSA} (\mathbf{Q}_{\texttt{A}}, \mathbf{K}_{\texttt{B}}, \mathbf{V}_{\texttt{B}})
    \end{aligned}
    \right.$ &  {$\mathcal{O}(N_{(\texttt{A})}^2)$}  &  \cite{lu2019vilbert, yun2021pano}  \\

Cross-Att. to Con. & 2-stream cross-att. followed by concat. & 2 $\rightarrow$ 1 & $\left\{ 
    \begin{aligned}
        \mathbf{Z}_{(\texttt{A})} & \gets {MHSA} (\mathbf{Q}_{\texttt{B}}, \mathbf{K}_{\texttt{A}}, \mathbf{V}_{\texttt{A}}) \\
        \mathbf{Z}_{(\texttt{B})} & \gets {MHSA} (\mathbf{Q}_{\texttt{A}}, \mathbf{K}_{\texttt{B}}, \mathbf{V}_{\texttt{B}}) \\
        \mathbf{Z} & \gets Tf(\mathcal{C}(\mathbf{Z}_{(\texttt{A})}, \mathbf{Z}_{(\texttt{B})}))
    \end{aligned}
    \right.$  & {$\mathcal{O}((N_{(\texttt{A})} + N_{(\texttt{B})})^{2})$} & \cite{hasan2021humor}   \cite{tsai2019multimodal,zhan2021product1m}  \\






\hline
\end{tabular}
}
\end{center}

\end{table*}


%% file: tex-2/section-4-applications.tex
\section{{Application Scenarios}}
\label{sec:applications-and-representative-models}


In this section we survey multimodal Transformers {based on the application scenarios.}
%
We consider two important paradigms:
(1) Transformers for multimodal pretraining (Section \ref{sec:transformers-for-multi-modal-pretraining}, including both task-agnostic (Section \ref{sec:task-agnostic-multi-modal-pretraining}) and task-specific (Section \ref{sec:task-specific-multi-modal-pretraining}) multimodal pretraining),
and
(2) Transformers for specific multimodal tasks (Section \ref{sec:transformers-for-specific-multi-modal tasks}).


\subsection{Transformers for Multimodal Pretraining}
\label{sec:transformers-for-multi-modal-pretraining}

Inspired by the great success of Transformer based pretraining in NLP community,
Transformers are also widely studied for multimodal pretraining as the various large-scale multimodal corpora is emerging.
Recent work has demonstrated that if pretrained on large scale multimodal corpora Transformer based models \cite{sun2019videobert, li2019visualbert, lu2019vilbert, su2019vl, chen2020uniter, zhou2020unified, tan2019lxmert} clearly outperform other competitors in a wide range of multimodal down-stream tasks, and moreover achieve the zero-shot generalization ability. 
These superiorities have led Transformer-based multimodal pretraining to become a hot topic, which 
has two main directions, \ie, general pretraining for agnostic down-stream tasks (Section \ref{sec:task-agnostic-multi-modal-pretraining}), goal-oriented pretraining for specific down-stream tasks (Section \ref{sec:task-specific-multi-modal-pretraining}).

We focus on these key points: 
(1) What trends are emerging? 
(2) Where/how do the cross-modal  interactions take place during pretraining? 
(3) How to sort out and understand the pretraining pretext objectives? 
How can they drive Transformers to learn the cross-modal interactions?


\subsubsection{Task-Agnostic Multimodal Pretraining}
\label{sec:task-agnostic-multi-modal-pretraining}

Recently Transformer-oriented pretraining has been widely studied involving diverse modality combinations, \eg, video-text \cite{sun2019videobert, sun2019learning, luo2020univl}, image-text \cite{lu2019vilbert,li2019visualbert,tan2019lxmert, yang2022vision, li2022grounded, zhang2022glipv2}, acoustic-text \cite{zheng2021fused}.

Among existing work,
the following main trends are emerging: 

{\textbf{(1)}} {Vision-language pretraining (VLP) is a major research problem in this field.} VLP is including both ``image + language'' and ``video + language'', also termed  visual-linguistic pretraining. A great deal of excellent work has been proposed, \eg, 
VideoBERT \cite{sun2019videobert},
ViLBERT \cite{lu2019vilbert},
LXMERT \cite{tan2019lxmert},
VisualBERT \cite{li2019visualbert},
VL-BERT \cite{su2019vl},
UNITER \cite{chen2020uniter},
CBT \cite{sun2019learning},
Unicoder-VL \cite{li2020unicoder},
B2T2 \cite{alberti2019fusion},
VLP \cite{zhou2020unified},
12-in-1 \cite{lu202012},
Oscar \cite{li2020oscar},
Pixel-BERT \cite{huang2020pixel},
ActBERT \cite{zhu2020actbert},
ImageBERT \cite{qi2020imagebert},
HERO \cite{li2020hero},
UniVL \cite{luo2020univl}, SemVLP \cite{li2021semvlp}. 

{\textbf{(2)}} {Speech can be used as text.} Thanks to  recent advances in  automatic speech recognition (ASR) techniques, in a multimodal context, speech can be converted to text by the
 off-the-shelf speech
recognition tools. 
For instance, VideoBERT \cite{sun2019videobert} and CBT \cite{sun2019learning} make full use of speech rather than low-level sounds as a source of cross-modal supervision, by extracting  high-level semantic text.

{\textbf{(3)}} {Overly dependent on the well-aligned multimodal data.} A majority of Transformer-based multimodal pretraining works in a self-supervised manner, however, it is overly dependent on the well-aligned multimodal sample pairs/tuples.
For instance, large amount of image-language pretraining Transformer models are pretrained on large-scale image-text pairs, \eg,
VisualBERT \cite{li2019visualbert}, VL-BERT \cite{su2019vl},  ViLBERT \cite{lu2019vilbert},
LXMERT \cite{tan2019lxmert}, UNITER \cite{chen2020uniter}.
For another example,
the instructional videos (\eg, cooking) \footnote{{Note that instructional videos also have weakly aligned cases \cite{miech2020end, han2022temporal}.}} are widely used as the pretraining corpora, \eg,
HowToVQA69M \cite{yang2021just},
HowTo100M \cite{miech2019howto100m},
as {in general,}
their visual clues/content and the spoken words have a higher probability to align with each other, {if compared with other videos.} 
However, using  cross-modal alignment as cross-modal supervision is costly for large-scale applications.
Thus, how to use the weakly-aligned or even unpaired/unaligned  multimodal data as the pretraining corpora is still understudied. Some recent attempts \cite{wang2021simvlm, zhan2021product1m} study the use of weakly-aligned cross-modal supervision  to train Transformers to learn the cross-modal interactions. 

{\textbf{(4)}} 
{Most of the existing pretext tasks transfer well across modalities.}
For instance, Masked Language Modelling (MLM) in the text domain has been applied to audio and image, \eg,  Masked Acoustic Modelling \cite{chen2020mam, zheng2021fused}, Masked Image Region Prediction \cite{murahari2020large},
while both Sentence Ordering Modelling (SOM) \cite{golestani2021using} in text domain and Frame Ordering Modelling (FOM) \cite{li2020hero} in video domain share the same idea.  
We will further discuss the pretext tasks for multimodal Transformer pretraining in the follows.

{\textbf{(5)}} {Model structures are mainly in three categories.} Essentially, in multimodal pretraining scenarios, Transformer models work based on those self-attention variants that are discussed in Section \ref{sec:self-attention-in-multimodal-context}. Thus, if considered from the perspective of model structures, the existing Transformers for multimodal pretraining are also mainly in three categories, \ie, single-stream, multi-stream, hybrid-stream. 

{\textbf{(6)}}
{Cross-modal interactions can perform within various components/levels in the pretraining pipelines.}
For  Transformer based multimodal pretraining, the key is to drive the Transformer (encoder w/, w/o decoder) to learn the cross-modal interactions.
In the existing Transformer-based multimodal pretraining practices, the cross-modal interactions are flexible, which can perform within various components/levels in the pretraining pipelines.
In general, Transformer-based multimodal pretraining pipelines have three key components, from bottom to top, \ie, tokenization, Transformer representation, objective supervision. 
For not only the multimodal pretraining but also the specific multimodal tasks,
the cross-modal interactions can perform within arbitrary component(s) of the three.
As discussed in Section \ref{sec:self-attention-in-multimodal-context},
because self-attention models the embedding of an arbitrary token from an arbitrary modality as a node of a graph,
 the existing pretraining pipelines
can, in general, be transferred independently across
modalities,
unless considered with modality-specific objectives.



\keypoint{Discussion} Vision Language Pretraining (VLP) follows two general pipelines: two-stage (need object detector, \eg, Faster R-CNN \cite{ren2015faster}) (\eg, LXMERT \cite{tan2019lxmert}, ViLBert \cite{lu2019vilbert}, VL-Bert \cite{su2019vl}, UNITER \cite{chen2020uniter}) and end-to-end (\eg, Pixel-Bert \cite{huang2020pixel}, SOHO \cite{huang2021seeing}, KD-VLP \cite{liu2021kd}, Simvlm \cite{wang2021simvlm}). 
Two-stage pipelines have a main advantage -- object-aware perceiving, by using the supervised pre-trained visual detectors, 
however these are based on a strong assumption that the visual representations can be fixed.


\keypoint{Discussion}
How to look for more corpora that intrinsically have well-aligned cross-modal supervision, such as instructional videos, is still an open problem.
However,
weakly-aligned cross-modal samples are popular in the real-life scenarios, for instance, enormous weakly aligned multimodal
data samples are emerging in e-commerce \cite{zhan2021product1m}, due to fine-grained categories, complex combinations, and fuzzy correspondence.
Well labelled/aligned cross-modal datasets are very costly in collecting and annotating;  
how to use weakly-aligned or even unaligned corpora crawled from the web is a promising question. 
Some recently successful practice \cite{radford2021learning, ramesh2021zero, wang2021simvlm}
used weakly aligned image-text pairs
to perform pretraining, and achieve both competitive performance and zero-shot learning capability for image classification, image-text retrieval, and open-ended visual question answering, \etc.
Because these practices in weak supervision  make full use of large-scale pretraining corpora, they yield greater promise of zero-shot generalization.






\keypoint{Pretext Tasks}
In Transformer based multimodal pretraining,
the pretraining tasks/objectives are also termed  pretext tasks/objectives.
To date, various pretext tasks have been studied, \eg, masked language modelling (MLM) \cite{zhan2021product1m},
masked image region prediction/{classification} (also termed  masked object classification (MOC)) \cite{murahari2020large, zhan2021product1m},
masked region regression (MRR) \cite{qi2020imagebert},
visual-linguistic matching (VLM) (\eg, image–text
matching (ITM) \cite{lin2020interbert}, image text matching (ITM), phrase-region alignment (PRA) \cite{liu2021kd}, word-region alignment (WRA) \cite{chen2020uniter}, video-subtitle matching (VSM) \cite{li2020hero}),
masked frame modelling (MFM) \cite{li2020hero},
frame order modelling (FOM) \cite{li2020hero},
next sentence prediction (NSP) \cite{devlin2018bert, lu2019vilbert, murahari2020large},
masked sentence generation (MSG) \cite{li2021scheduled},
masked group modelling (MGM) \cite{lin2020interbert}, 
prefix language modelling (PrefixLM) \cite{wang2021simvlm},
video conditioned masked
language model \cite{luo2020univl},
text conditioned masked frame
model \cite{luo2020univl},
visual translation language modelling
(VTLM) \cite{zhou2021uc2}, and
image-conditioned masked language modelling (also termed  image-attended masked language modelling) \cite{hao2020towards}.
 These down-stream task -agnostic pretext pretraining is optional,
and the down-stream task objectives can be trained directly,
which will be discussed in Section~\ref{sec:task-specific-multi-modal-pretraining}.
{Table~\ref{table:pretext-tasks} provides the common and representative pretext tasks for Transformer based multimodal pretraining.}
\blue{In practice, pretext tasks can be combined, and some representative cases are summarized in Table 3 of \cite{ruan2021survey}, Table 2 of \cite{chen2022vlp}.}

\input{tables/pretext-tasks}

The pretext tasks have multiple taxonomies: 

{\textbf{(1)}} {Supervision.}
{The common multimodal pretraining Transformers use well-aligned,  weakly-aligned, and even unaligned multimodal sample pairs/tuples, to work in supervised, weakly-supervised, and unsupervised manners, respectively.}
Meanwhile, 
if we consider the definitions of their pretext tasks/objectives from supervision, the pretexts can be sorted into unsupervised/self-supervised (\eg, masked language modelling (MLM) \cite{sun2019videobert, zhan2021product1m})  and supervised (\eg, image-text matching (ITM)    \cite{lin2020interbert} \cite{chen2020uniter,  li2019visualbert,  lu2019vilbert,  tan2019lxmert, yao2021filip}), \etc.
Nowadays, self-supervised attempts are the majority. 

{\textbf{(2)}} {Modality.} Considering the mathematical formulations, some pretexts are defined on single modality, \eg, masked language modelling \cite{sun2019videobert}, masked acoustic modelling \cite{chen2020mam}, masked region regression (MRR) \cite{qi2020imagebert},
while other pretexts are defined on multiple modalities, \eg, image-conditioned masked language modelling (IMLM) \cite{xia2021xgpt}, image-text matching (ITM) \cite{lin2020interbert}, video-subtitle matching (VSM) \cite{li2020hero}.
Thus, from this mathematical view, the pretext tasks can be  divided into two categories, \ie, uni-modal and multimodal.  

However, this classification is not really accurate.
It should be highlighted that in multimodal pretraining Transformer models, even if the pretext objective formulations only include uni-modal elements, pretexts can still involve other modalities, essentially conditioned on the clues from other modalities, by (a) prepositive token level interactions and/or Transformer level interactions, 
(b) co-training with other pretexts that involve other modalities. 
For instance, VL-BERT \cite{su2019vl} uses two dual pretext tasks, \ie, masked language modelling and masked RoI classification.

{\textbf{(3)}} {Motivation.} If consider their motivations, the pretext tasks include masking, describing, matching, ordering, \etc.




Some recent surveys focus on VLP and compare the existing VLP Transformer models from the angles of domain (image-text or video-text), vision feature extraction, language feature extraction, architecture (single- or dual- stream), decoder (w/, w/o), pretext tasks/objectives, pretraining datasets, and down-stream tasks, \eg, Table 3 of \cite{ruan2021survey}, Table 2 of \cite{chen2022vlp}.
Different from these views,
in this survey,
we would propose our comparisons from some new perspectives.
Specifically:
{\textbf{(1)}}
The core of Transformer ecosystem is self-attention, thus we would compare the existing multimodal pretraining Transformer models from the angles of how and when the self-attention or its variants perform cross-modal interactions. 
{\textbf{(2)}} Considering from a geometrically topological perspective, self-attention helps Transformers intrinsically 
work in a modality agnostic pipeline that is compatible
with various modalities by taking in the embedding of each
token as a node of graph, thus we would highlight that the existing VLP can be applied to other modalities, beyond visual and linguistic domains. 
{\textbf{(3)}} 
We suggest to treat the Transformer-based multimodal pretraining pipelines having three key components, from bottom to top, \ie, tokenization, Transformer representation, objective supervision.



\keypoint{Discussion}
In spite of the recent advances,
multimodal pretraining Transformer methods still have some obvious bottlenecks.
\magenta{For instance,
as discussed by \cite{xia2021xgpt} in VLP field,
while the BERT-style cross-modal pretraining models produce excellent results on various down-stream vision-language tasks, they fail to be applied to generative tasks directly.}
As discussed in \cite{xia2021xgpt},
both VideoBERT \cite{sun2019videobert} and CBT \cite{sun2019learning} have to train a separate video-to-text decoder for video captioning.
This is a significant gap between the pretraining models designed for discriminative and generative tasks, as the main reason is discriminative task oriented pretraining models do not involve the decoders of Transformer.
Therefore, how to design more unified pipelines that can work for both discriminative and generative down-stream tasks is also an open problem to be solved.
Again for instance,
common multimodal pretraining models often underperform for fine-grained/instance-level tasks as discussed by \cite{zhan2021product1m}.


\keypoint{Discussion}
\magenta{As discussed in \cite{xia2021xgpt},
the masked language
and region modelling as pre-training task have a main advantage that the Transformer encoder learned from these supervisions can encode both vision and language patterns based on bidirectional context and it is naturally fit for the semantic understanding tasks, \eg, VQA, image-text retrieval.}


\keypoint{Discussion}
How to boost the performance for multimodal pretraining Transformers is an open problem.
Some practices demonstrate that 
multi-task training (by adding auxiliary loss) \cite{zhan2021product1m, lu202012} and adversarial training \cite{gan2020large} improve multimodal pretraining Transformers to further boost the performance. 
Meanwhile, overly compound
pretraining objectives potentially upgrade the challenge of balancing among different loss terms, thus complicate the training optimization \cite{wang2021simvlm}.
Moreover, 
the difficulty of the pretexts is also worth discussing.
In general, if aim to learn more explicit object concepts, more complex pretext losses will be used \cite{liu2021kd}. However, for pretexts, whether more complexity is better remains a question.





\subsubsection{Task-Specific Multimodal Pretraining}
\label{sec:task-specific-multi-modal-pretraining}

In practices of multimodal Transformers,
the aforementioned down-stream task -agnostic pretraining is optional, not necessary,
and down-stream task specific pretraining is also widely studied \cite{ zhang2021ernie, guhur2021airbert, xia2021xgpt, murahari2020large}.
The main reasons include:
(1) Limited by the existing  technique, it is extremely difficult to design a set of highly universal network architectures, pretext tasks, and corpora that work for all the various down-stream applications.
(2) There are non-negligible gaps among various down-stream applications, \eg, task logic, data form, making it difficult to transfer from pretraining to down-stream applications.

Therefore,
a large number of down-stream tasks still need tailor-made pretraining to improve the performance.
Guhur \etal \cite{guhur2021airbert} propose in-domain pretraining for vision-and-language navigation, as 
the general VLP
focuses on learning vision-language correlations, not designed for sequential decision making as required
in embodied VLN.
Murahari \etal \cite{murahari2020large} present a visual dialogue oriented approach to leverage pretraining on general
vision-language datasets. 
XGPT \cite{xia2021xgpt} is tailor-made for image captioning, to overcome the limitation that BERT-based cross-modal pre-trained models fail to be applied
to generative tasks directly.
ERNIE-ViLG \cite{zhang2021ernie} is designed for bidirectional image-text generation with Transformers.



Special modalities have their own unique domain knowledge that can be used to design the specific pretrain pretexts.
GraphCodeBERT \cite{guo2020graphcodebert} uses
two structure-aware pretext tasks (\ie, predict where a variable is identified from, data flow edge
prediction between variables) for programming source code.
To learn from the spatial cues in $360^{\circ}$ video,
Morgado \etal \cite{morgado2020learning} propose to perform contrastive audio-visual spatial alignment of $360^{\circ}$ video and spatial audio.
Med-BERT \cite{rasmy2021med} is a contextualized embedding model pretrained on a structured electronic health record dataset of two million patients. 
Kaleido-BERT \cite{zhuge2021kaleido} is a VLP Transformer model tailor-made for the fashion domain.

\subsection{Transformers for Specific Multimodal Tasks}
\label{sec:transformers-for-specific-multi-modal tasks}


Recent work has demonstrated that Transformer models can encode various multimodal inputs in both classical and novel discriminative applications, \eg, RGB \& optical flow \cite{gavrilyuk2020actor}, 
{RGB \& depth \cite{wang2022multimodal},
RGB \& point \blue{cloud} \cite{wang2022bridged},
RGB \& LiDAR \cite{prakash2021multi, bai2022transfusion},}
textual description \& point cloud \cite{zhao20213dvg}, acoustic \& text \cite{zheng2021fused}, audio \& visual observation for Audio-Visual Navigation \cite{chen2021semantic},
speech query \& schema of SQL database \cite{song2022speech},
text question/query \& the schema SQL database \cite{cai2021sadga},
audio \& tags \cite{favory2021learning},
multimodal representation for video  \cite{gabeur2020multi, Shvetsova_2022_CVPRMulti},
text query \& video \cite{wang2022multi},
audio \& video for audio visual speech enhancement (AVSE) \cite{ramesh2021vset},
audio \& video for Audio-Visual Video Parsing \cite{lin2021exploring},
audio \& video for audio-visual speech recognition \cite{afouras2018deep},
video \& text for Referring Video Object Segmentation (RVOS) \cite{botach2021end},
source code \& comment \& data flow \cite{guo2020graphcodebert},
image \& text for retrieval \cite{hong2021gilbert}.

Meanwhile, Transformers also contribute to various multimodal generative tasks, including single-modality to single-modality (\eg, raw audio to 3D mesh sequence \cite{fan2021faceformer},
RGB to 3D scene \cite{shin20193d},
single image to 3D human texture estimation \cite{xu20213d},
RGB to scene graph \cite{lin2020gps, guo2021general, wang2021topic, lu2021context},
graph to graph \cite{ammanabrolu2021learning},
knowledge graph to text \cite{ke2021jointgt},
video to scene graph \cite{teng2021target},
video to caption \cite{chen2018tvt, lin2021swinbert, deng2021sketch, wang2021end},
image to caption  \cite{huang2019attention, pan2020x, yang2021causal,luo2021dual, xu2021towards},
text to speech \cite{li2019neural},
text to image \cite{ ramesh2021zero, ding2021cogview},
text to shape \cite{sanghi2021clip},
RGB to 3D human pose and mesh \cite{lin2021end},
music to dance \cite{huang2020dance}),
multimodality to single modality (\eg, 
image \& text to scene graph \cite{zhong2021learning},
Video Dialogue (text \& audio \& visual to text) \cite{geng2021dynamic},
Mono Audio \& Depth to Binaural Audio \cite{parida2022beyond},
music piece \& seed 3D motion to long-range future 3D motions  \cite{li2021ai},
X-raying image \& question to answer \cite{jaunet2021visqa},
video \& text \& audio to text \cite{lin2021vx2text}),
and multimodality to multimodality (\eg, \cite{lin2021m6}).

%% file: tables/pretext-tasks.tex
\begin{table*}[!t]
\caption{ Pretext task comparison of multi-modal pretraining Transformer models (for agnostic down-stream tasks). 
``C-M Loss'': cross-modal loss;
``Con. Loss'': loss conditioned on other modality/modalities.
}
\label{table:pretext-tasks}
\begin{center}
\resizebox{\textwidth}{!}{
\begin{tabular}{ c | c | c | c | c }
\hline
Types (Motivations) & Tasks & C-M Loss & Con. Loss  & References \\
\hline
\hline
\multirow{10}{*}{Masking} & Masked Language Modelling (MLM) &   & $\checkmark$  &  \cite{sun2019videobert}, \cite{zhan2021product1m} \\

 &  Image-Conditioned Masked Language Modelling (IMLM)  & & $\checkmark$ &   \cite{hao2020towards, zhou2021uc2} \cite{xia2021xgpt} \\

   & Text-Conditioned Masked Region Prediction  & & $\checkmark$ &  \cite{zhou2021uc2} \\

  & Masked Acoustic Modelling   & & $\checkmark$ &  \cite{chen2020mam, zheng2021fused} \\

  &  Masked Image Region Regression   & & $\checkmark$  &   \cite{qi2020imagebert} \\

    & Masked Image Region Prediction  & & $\checkmark$ &   \cite{murahari2020large} \\

      & Masked Frame Modelling (MFM)  & &  $\checkmark$ & \cite{li2020hero} \\

         & Masked Sentence Generation (MSG)  &  & $\checkmark$ &    \cite{li2021scheduled} \\

            &  Video Conditioned Masked
Language Model  & & $\checkmark$ &  \cite{luo2020univl}\\

 & Text Conditioned Masked Frame
Model   &  & $\checkmark$ &    \cite{luo2020univl}\\
\hline
\hline
\multirow{3}{*}{Describing}  & Image-conditioned Denoising Autoencoding (IDA)  & &  $\checkmark$ &   \cite{xia2021xgpt} \\
    & Text-conditioned Image Feature Generation (TIFG)  & & $\checkmark$ &  \cite{xia2021xgpt} \\

      & Prefix Language Modelling (PrefixLM)  & $\checkmark$ &  &  \cite{wang2021simvlm}  \\
      
\hline
\hline
\multirow{5}{*}{Matching} & Image-Text Matching (ITM)  & $\checkmark$ & &   \cite{lin2020interbert} \cite{chen2020uniter,  li2019visualbert,  lu2019vilbert,  tan2019lxmert, yao2021filip}, 
\\
 & Phrase-Region Alignment (PRA)  & $\checkmark$ & &  \cite{liu2021kd} \\
  & Word-Region Alignment (WRA)  & $\checkmark$ & &  \cite{chen2020uniter},  \cite{kim2021vilt} \\
   & Video-Subtitle Matching (VSM)  & $\checkmark$ &  & \cite{li2020hero} \\
    & Next Sentence Prediction (NSP)  & & $\checkmark$ &  \cite{devlin2018bert, lu2019vilbert, murahari2020large} \\
\hline
\hline
\multirow{2}{*}{Ordering} & Sentence Ordering Modelling (SOM)  & & $\checkmark$ &  \cite{golestani2021using} \\
 & Frame Ordering Modelling (FOM)  & & $\checkmark$ &  \cite{li2020hero} \\

\hline
\end{tabular}
}
\end{center}

\end{table*}


%% file: tex-2/section-5-challenges.tex
\section{Challenges and Designs}
\label{sec:challenges-and-designs}
Complementing the application {scenario taxonomy} discussed in Section \ref{sec:applications-and-representative-models},
we further survey prior work from the perspective of technical challenges.
We discuss seven challenges of Transformer based multimodal learning, including fusion, alignment, transferability,
efficiency,
robustness,
universalness, and interpretability.
This further extends the taxonomy introduced in \cite{baltruvsaitis2018multimodal} to tackle the higher diversity and wider scopes of 
existing Transformer based MML works in recent years.




\subsection{Fusion}
In general, MML Transformers fuse  information across 
multiple modalities primarily at three levels:
input (\ie, early fusion), intermediate representation (\ie, middle fusion), and prediction (\ie, late fusion).
%
%
Common early fusion based MML Transformer models \cite{sun2019videobert,li2019visualbert,li2020unicoder}
 are also known as {one-stream architecture},
allowing the adoption of the merits of BERT
due to minimal architectural modification.
The main difference between  these one-stream models
is the usage of problem-specific modalities with variant masking techniques.
With attention operation, a noticeable fusion scheme
is introduced based on a notion of bottleneck tokens \cite{nagrani2021attention}.
It applies for both early and middle fusion
by simply choosing to-be-fused layers.
We note that the simple prediction-based late fusion \cite{chen1998audio,owens2018audio} is less adopted 
in MML Transformers.
This makes sense considering the motivations of learning stronger multimodal contextual representations and great advance of computing power. 
For enhancing and interpreting the fusion of MML,
probing the interaction and measuring the fusion between modalities \cite{xue2021probing}
would be an interesting direction to explore.













\subsection{Alignment}
Cross-modal alignment is the key to 
a number of real-world multimodal applications.
Transformer based cross-modal alignment has been studied for various tasks, \eg,
speaker localization in multi-speaker videos \cite{truong2021right},
speech translation \cite{zheng2021fused},
text-to-speech alignment \cite{chen2020multispeech},
text-to-video retrieval \cite{ging2020coot, patrick2020support, gabeur2022masking},
and visual grounding of natural language \cite{sadhu2020video, zhang2021explainable, chen2021end, Chen_2022_CVPR, Yang_2022_CVPR_TubeDETR}.
Recently, Transformer based alignment
\cite{radford2021learning,jia2021scaling,xu2021videoclip,lei2021less, yang2021taco}
has led to a surge of leveraging large quantities of web data (\eg, image-text pairs) for vision and language tasks.

A representative practice is to map two modalities into a common representation space with contrastive learning over paired samples.
The models based on this idea are often enormous in size
and expensive to optimize from millions or billions of training data.
Consequently, successive works mostly exploit pretrained models
for tackling various down-stream tasks \cite{li2021align,wang2022clip,luo2021clip4clip,fang2021clip2video,narasimhan2021clip}.
These alignment models
have the ability of zero-shot transfer
particularly for image classification via \blue{prompt engineering~\cite{liu2023pre}}.
This novel perspective is mind-blowing, given that image classification is conventionally regarded as a unimodal learning problem
and zero-shot classification remains an unsolved challenge despite extensive research \cite{xian2018zero}.
This has been  studied
for more challenging and fine-grained tasks (\eg, object detection \cite{gu2021open}, visual question answering \cite{li2021align,chen2020uniter,li2020oscar,tan2019lxmert}, and instance retrieval \cite{li2021align,hong2021gilbert})
by 
imposing region (semantic parts such as objects) level alignment.
Fine-grained alignment will however incur more  computational costs
from explicit region detection 
and how to eliminate this whilst keeping the region-level learning capability becomes a challenge. 
Several ideas introduced recently include
random sampling \cite{huang2020pixel},
learning concept dictionary \cite{huang2021seeing},
uniform masking \cite{cho2020x},
patch projection \cite{kim2021vilt},
joint learning of a region detector \cite{xu2021e2e}, and
representation aligning before mask prediction \cite{li2021align}.

\subsection{Transferability}
\label{sec:transferability}




Transferability is a major challenge for Transformer based multimodal learning, involving the question of how to transfer models across different datasets and applications.

Data augmentation and adversarial perturbation strategies help multimodal Transformers to improve the generalization ability. VILLA \cite{gan2020large} is a two-stage  strategy (task-agnostic adversarial pretraining, followed by task-specific adversarial finetuning) that improves VLP Transformers.

In practice,
the distribution gap between training data and practical data is noticeable.
For instance,
supervised data samples (well-labelled, well-aligned) are costly in practical applications, thus
how to transfer the supervised multimodal Transformers pretrained on well-aligned cross-modal pairs/tuples to the weakly aligned test bed is challenging \cite{zhan2021product1m}.  
CLIP \cite{radford2021learning} is an inspiring solution that transfers knowledge across modalities by learning a shared multimodal embedding space, enabling zero-shot transfer
of the model to down-stream tasks.
The main inspiration that CLIP presents the community is that the pretrained multimodal (image and text) knowledge can be transferred to down-stream zero-shot image prediction by using a prompt template ``\texttt{A photo of a \{label\}.}'' to bridge the distribution gap between training and test datasets.


Over-fitting is a major obstacle to transfer. Multimodal Transformers can be overly fitted to the dataset
biases during training, due to the large modelling capability.
Some recent practices exploit how to transfer the oracle model trained on noiseless dataset to real dataset.
For instance, 
Kervadec \etal \cite{kervadec2021supervising, kervadec2021transferable} explore 
how transferable reasoning patterns are in VQA,
and demonstrate that for LXMERT \cite{tan2019lxmert}/BERT-like
reasoning patterns can be partially transferred from an ideal dataset to a real dataset.

Cross-task gap is another major obstacle to transfer \cite{rahman2020integrating, xia2021xgpt}, due to the different reasoning and input-output workflows, \eg,
how to use multimodal datasets to finetune the language pretrained model is difficult \cite{rahman2020integrating}.
In real applications, 
multimodal pretrained Transformers sometimes need to handle the uni-modal data at inference stage due to the issue of missing modalities. One solution is using knowledge
distillation, \eg, distilling
from multimodal to uni-modal attention in Transformers \cite{agarwal2021multimodal}, distilling from multiple uni-modal Transformer teachers to a shared Transformer encoder \cite{li2021towards}.
There is a huge gap across discriminative and generative multimodal tasks.
As discussed in \cite{xia2021xgpt},
the BERT-like encoder-only multimodal Transformers (\eg, VideoBERT \cite{sun2019videobert}, CBT \cite{sun2019learning}) need
separately to train decoders for generation tasks. This could create a pretrain-finetune discrepancy detrimental to the
generality.
Recently, more and more attempts study this issue further, \eg,
GilBERT \cite{hong2021gilbert} is a generative VLP models for
a discriminative task, \ie, image-text retrieval.

Cross-lingual gap also should be considered for the transferability of Transformer based multimodal learning, \eg,
universal cross-lingual generalization from English to non-English multimodal contexts \cite{zhou2021uc2,ni2021m3p}.









\subsection{Efficiency}
\label{sec:efficiency}
Multimodal Transformers suffer from two major efficiency issues:
(1) Due to the large model parameter capacity, they are data hungry and thus dependent  on huge scale training datasets.
(2) They are limited by the time and memory complexities that grow quadratically with the input sequence length, which are caused by the self-attention.
In multimodal contexts, 
calculation explosion will become worse due to jointly high dimension representations.
These two bottlenecks are interdependent and should be considered together.

To improve the training and/or inferring efficiency for multimodal Transformers,
 recent efforts have attempted to find various solutions, to use fewer training data and/or parameters.
The main ideas can be summarized as the follows. 

{\textbf{(1)}} {Knowledge distillation.}
Distill the knowledge from the trained larger Transformers to smaller Transformers \cite{touvron2021training}.
Miech \etal \cite{miech2021thinking} conduct distillation from a slower model (early concatenation based Transformers, $\mathcal{O}((N_{(\texttt{A})} + N_{(\texttt{B})})^{2})$) to a faster one (independently dual branch Transformers, $\mathcal{O}(N_{(\texttt{A})}^{2})$).

{\textbf{(2)}} {Simplifying and compressing model.}
Remove the components to simplify the pipelines.
Taking the VLP Transformer models as an example,
two-stage pipeline is costly as they need object detector.
One simplifying is processing the visual input in convolution-free manner, \eg, E2E-VLP \cite{xu2021e2e},  ViLT  \cite{kim2021vilt}. 
DropToken \cite{akbari2021vatt}  reduces
the training complexity via
random dropping a portion of the video and audio tokens from input sequence during training.
DropToken can be treated as an implementation of dropout or adversarial training.
Weight-sharing is also a common practice for simplifying multimodal Transformer models.
Wen \etal \cite{wen2021cookie} present a weight-sharing Transformer on top of the visual and
textual encoders to align text and image.
Lee \etal \cite{lee2020parameter} propose a novel parameter sharing scheme based on low-rank approximation.

{\textbf{(3)}} {Asymmetrical network structures.}
Assign different model capacities and computational size properly for different modalities, to save parameters. See Figure 2 in \cite{kim2021vilt}.

{\textbf{(4)}} {Improving utilization of training samples.} 
Liu \etal \cite{liu2021multi} train a simplified LXMERT by making full use of fewer samples at different granularities.
Li \etal \cite{li2021supervision} use fewer data to train CLIP by  fully mining the  potential self-supervised signals of (a) self-supervision within each modality, (b) multi-view supervision across modalities, and (c) nearest-neighbour  supervision from other similar pairs.

{\textbf{(5)}} {Compressing and pruning model.}
Search the optimal sub-structures/sub-networks of multimodal Transformers, \eg,
playing Lottery Tickets with the VLP Transformer models \cite{gan2021playing}, adaptively freezing some layers during training \cite{he2021pipetransformer}.

{\textbf{(6)}} {Optimizing the complexity of self-attention.}
Transformers cost time and memory that grows quadratically with the input sequence length~\cite{dao2022flashattention}.
One potential solution is optimizing the $\mathcal{O}(N^{2})$ complexity, \eg,
Child \etal \cite{child2019generating} present sparse factorizations of the attention matrix to reduce the quadratical complexity to $\mathcal{O}(n \sqrt{n})$,
Transformer-LS \cite{zhu2021long} is an efficient Transformer for both language and vision long sequence, with linear computational and memory complexity.

{\textbf{(7)}} {Optimizing the complexity of self-attention based multimodal interaction/fusion.}
Nagrani \etal \cite{nagrani2021attention}  propose Fusion via Attention Bottlenecks  (FSN, fusion bottleneck) to improve the early concatenation based multimodal interaction.
FSN passes on the messages through a small number of bottleneck 
latents, thus requiring the model to purify the most necessary information
from each modality for cross-modal sharing.
This strategy uses the fusion bottleneck as a bridge, and not only
improves fusion performance, but also reduces computational cost.

{\textbf{(8)}} {Optimizing other strategies.}
Use optimal strategies to perform the common Transformer based multimodal interactions.
Given the quadratic complexity of self-attention,
using early concatenation based multimodal interaction to synchronously fuse the inputs from multiple modalities/views is costly.
Yan \etal \cite{yan2022multiview} present an efficient solution that sequentially fuses information between all pairs of
two adjacent views in ascending order of sequence length.
This is intrinsically a greedy strategy.

\subsection{Robustness}
\label{sec:robustness}
Multimodal Transformers pretrained on large-scale corpora achieve the state-of-the-art for various multimodal applications, while their robustness is still unclear and understudied.
This at least involves two key challenges, \ie, how to theoretically analyse the robustness, how to improve the robustness.

Although that recent attempts \cite{wang2020rethinking, paul2021vision, mao2021towards, Ma_2022_CVPR_are_multimodal} study and evaluate how the Transformer components/sub-layers contribute to the robustness, 
the main bottleneck is that the community lacks theoretical tools to analyse the Transformer family. 
Recently, the common practices to analyse robustness are mainly based on experiment evaluations \cite{akula2021robust}, \eg, cross-dataset evaluations, perturbation-based evaluations.
Thus, some multimodal datasets   \cite{akula2020words, li2021adversarial} are proposed for evaluating the robustness.

Recent attempts mainly use two straightforward methods to improve the robustness for multimodal Transformer models: (1) augmentation and adversarial learning based strategies \cite{li2020closer, zhang2021domain},
(2) fine-grained loss functions \cite{kant2021contrast}.
For instance:
VILLA \cite{gan2020large}  is a generic adversarial training framework that can be applied to various multimodal Transformers.
Akula \etal \cite{akula2020words} empirically demonstrate that
ViLBERT fails to exploit linguistic structure, and they propose two methods to improve the robustness of ViLBERT, one based on contrastive learning and the other based on multi-task learning.

\subsection{Universalness} 
\label{sec:universalness}

Due to the highly diversity of tasks and modalities of multimodal learning, universalness is an important problem for multimodal Transformer models.
A large amount of recent attempts \cite{
pramanik2019omninet, luo2020univl, 
wang2022unifying, girdhar2022omnivore} study how to use as unified as possible pipelines to handle various modalities and multimodal tasks.
Ideally, the unified multimodal Transformers can be compatible with various data (\eg,
aligned and unaligned, uni-modal and multimodal) and tasks (\eg, supervised and unsupervised, uni-modal and multimodal, discriminative and generative), and meanwhile have either few-shot or even zero-shot generalization ability. Thus, the current solutions for universalness goal for multimodal Transformers are preliminary probes.

The currently unifying-oriented attempts mainly include: 

{\textbf{(1)}} {Unifying the pipelines for both uni-modal and multimodal inputs/tasks.} 
As discussed Section \ref{sec:transferability}, in practical scenarios,
multimodal Transformers need to
handle  uni-modal data due to the issue of missing modalities.
Distilling multimodal knowledge into  small models that are adaptable to uni-modal data and tasks is a successful practice \cite{agarwal2021multimodal, li2021towards}.

{\textbf{(2)}} {Unifying the pipelines for both  multimodal understanding and generation.}
In general, for multimodal Transformer pipelines,
understanding and discriminative tasks require Transformer encoders only, while generation/generative tasks require both Transformer encoders and decoders.
Existing attempts use multi-task learning to combine the understanding and generation workflows, where two kinds of workflows are jointly trained by multi-task loss functions.
From the perspective of model structures, typical solutions include:
(a) encoder + decoder, \eg, E2E-VLP \cite{xu2021e2e}.
(b) separate encoders + cross encoder + decoder, \eg, UniVL \cite{luo2020univl}, CBT \cite{sun2019learning}.
(c) single unified/combined encoder-decoder, \eg, VLP \cite{zhou2020unified}.
(d) \blue{two-stream} decoupled design \cite{li2021scheduled}. 

{\textbf{(3)}} {Unifying and converting the tasks themselves}, \eg, CLIP \cite{radford2021learning} converts zero-shot recognition to retrieval, thus reduces the costs of modifying the model.

However, the aforementioned practices suffer some obvious challenges and  bottlenecks, at least including: 

{\textbf{(1)}} Due to modality and task gaps, universal models should consider the trade-off between universalness and cost. Unifying the pipelines of different modalities and tasks generally cause larger or more complicated model configuration, whereas for a specific modality or task, some components are redundant. 

{\textbf{(2)}} Multi-task loss functions increase the complexity of training. How to co-train multiple objectives properly  and effectively is challenging, due to that different objectives generally should be optimized in different strategies. 

\subsection{Interpretability}
\label{sec:interpretability}

Why and how  Transformers perform so well in multimodal learning has been investigated \cite{cao2020behind,chen2020uniter, hendricks2021decoupling,hendricks2021probing,frank2021vision, chefer2021generic, parcalabescu2021valse, zhao2022vl, Aflalo_2022_CVPR}.
These attempts mainly use probing task and ablation study. 
Cao \etal \cite{cao2020behind} design a set of probing tasks on UNITER \cite{chen2020uniter} and LXMERT \cite{tan2019lxmert}, to evaluate what patterns are learned in pretraining.
Hendricks \etal \cite{hendricks2021probing} probe the  image–language Transformers by fine-grained image–sentence pairs, and find that verb understanding is harder than subject or object understanding.
\magenta{Chen \etal \cite{chen2020uniter} examine the optimal combination of pretraining
tasks via ablation study, to compare how different pretexts contribute to the Transformers.}
Despite these attempts,
the interpretability of multimodal Transformers is still under-studied to date.

%% file: tex-2/section-6-discussion.tex
\section{Discussion and Outlook}
\label{sec:discussion-and-outlook}


Designing the universal MML models  to excel across all the unimodal and multimodal down-stream tasks with different characteristics simultaneously \cite{qi2020imagebert,cao2020behind} is a non-trivial challenge.
For instance, two-stream architectures \cite{radford2021learning,li2021align} are typically preferred over one-stream ones 
for cross-modal retrieval-like tasks in efficiency, since the representation of each modality can be pre-computed beforehand and reused repeatedly. 
That being said, how to design task-agnostic MML architectures is still an open challenge, in addition to other design choices such as pretext and objective loss functions.
Furthermore, a clear gap remains
between the state-of-the-art and this ultimate goal.
In general, existing multimodal Transformer models \cite{wang2021simvlm,li2021align,radford2021learning}
are superior only for specific MML tasks,
%
as they are designed specifically for only a subset of specific tasks
\cite{zhan2021product1m,zhuge2021kaleido,fang2021clip2video,hu2021scaling,narasimhan2021clip,lei2021less,xu2021videoclip,xue2021probing}.
%
%
Encouragingly, several recent studies towards { universal modality learning} in terms of modality-agnostic network design \cite{jaegle2021perceiver} and more task-generic architecture design \cite{mu2021slip,xu2021vlm,li2022blip} have been introduced,
and it is hoped this will spark further investigation.
To that end, instead of exhaustively exploring the vast model design space, {seeking in-depth understanding and interpretation of a MML model's behaviour might be insightful for superior algorithm design}, even though the interactions and synergy across different modalities are intrinsically complex and even potentially inconsistent over tasks
\cite{xue2021probing}.

For more fine-grained MML, it is widely acknowledged that discovering the latent semantic alignments across modalities is critical.
An intuitive strategy is to leverage semantic parts (\eg, objects) pre-extracted by an off-the-shelf detector for MML \cite{zhang2021vinvl,chen2020uniter,li2020oscar,tan2019lxmert,li2019visualbert,su2019vl,liu2021kd}.
This, however, is not only complex and error-prone, but computationally costly \cite{hao2020towards}.
Several remedies introduced recently include
random sampling \cite{huang2020pixel},
learning concept dictionary \cite{huang2021seeing},
jointly learning a region detector \cite{xu2021e2e}, and
representation aligning before mask prediction \cite{li2021align}.
Given the scale of MML training data, exploring this direction
needs exhaustive computational costs, and
it is supposed that industrial research teams with rich resources are more likely to afford. 
{Ideally, a favourable MML method would leave fine-grained
semantic alignment across modalities to emerge on its own},
which is worthy of careful investigation in the future.

As the learning scale expands exponentially,
the training data become inevitably noisy and heterogeneous \cite{radford2021learning,li2021align,wang2021simvlm}.
It has been recently shown that  properly tackling  the noise issue is useful \cite{li2022blip,li2021align}.
Another related facet is training strategy,
\eg, how many stages of training is superior over the common one-stage policy \cite{qi2020imagebert}.
%
Further, the quadratic complexity with Transformers
becomes more acute for multimodal data due to longer input.
Despite extensive research on efficient variants \cite{tay2020efficient},
{dedicated efficiency study for MML is still underestimated even empirically and call for more investigation}.
Identifying the strengths of Transformers for multimodal machine learning is a big open problem.
The following main points can be summarized from the literature:
{\textbf{(1)}} Transformers can encode implicit knowledge \cite{marino2021krisp}.
{\textbf{(2)}} The multi-head brings multiple modelling sub-spaces that can further enhance the expressive ability of the model.
Ideally, multiple heads after training are good and different.
This is essentially a good practice of ensemble learning.
{\textbf{(3)}} Transformers intrinsically have a nature of global aggregation that perceives the non-local patterns. 
{\textbf{(4)}} Thanks to the large model capacity, Transformer models handle the challenging domain gaps and shifts (\eg, linguistic and visual) better via effective pretraining on large-scale corpora \cite{zhang2021domain}.
{\textbf{(5)}} Transformers can represent the inputs as graphs, which are
intrinsically compatible with more modalities, \eg, table and SQL.
{\textbf{(6)}} For modelling series and sequence patterns (\eg, time-series), Transformers have better training and inference efficiency against RNN-based models,
thanks to their parallel computation in training and/or inference. 
Transformers are inherently permutation invariant for processing a sequence of points, \eg, well-suited for point cloud learning \cite{guo2021pct}.
{\textbf{(7)}} Tokenization makes Transformers flexible to organize  multimodal inputs, as discussed in Section \ref{sec:tokenized-input}.

%% file: tex-2/section-7-conclusion.tex
\section{Conclusion}
\label{sec:conclusion}
This survey focuses on  multimodal machine learning with Transformers.
We reviewed the landscape by introducing the Transformer designs and training in the multimodal contexts.
We summarized the key challenges and solutions for this emerging and exciting field.
Moreover, we discussed open problems and potential research directions. We hope that this survey gives a helpful and detailed overview for
new researchers and practitioners, provides
a convenient reference for relevant experts (\eg, multimodal machine learning researchers, Transformer network designers), and encourages
future progress.

%% file: tex-2/appendix.tex
\keypoint{{Notations and Abbreviations}}
Throughout this survey,
unless specified otherwise, mathematical symbols and abbreviated terms follow the conventions in Table \ref{table:definitions}.

\input{tables/definition_table}

\keypoint{{Special/Customized Tokens}}
{In both uni-modal and multimodal Transformers,
various special/customized tokens are semantically defined as place-holders in the token sequences.
Some common special tokens are summarized in Table~\ref{table:tokens}.}



\input{tables/tokens}

%% file: tables/definition_table.tex
\begin{table}[!h]
\caption{Notation and abbreviations used in this survey.}
\label{table:definitions}
\begin{center}
\resizebox{\columnwidth}{!}{
\begin{tabular}{ l | l }
\hline
Notations & Descriptions \\
\hline
$\mathbf{M}$, $\mathbf{M}^{T}$ & matrix ${\bf M}$ and its transpose \\
$\mathbf{Q}$ & Query matrix of Self-Attention \\
$\mathbf{K}$ & Key matrix of Self-Attention \\
$\mathbf{V}$ & Value matrix of Self-Attention \\
$SA(\cdots)$ & Self-Attention \\
$MSA(\cdots)$ & Masked Self-Attention \\
$FFN(\cdot)$ & Feed-Forward Network \\
$MHSA(\cdots)$ & Multi-Head Self-Attention \\
$MLP(\cdots)$ & Multi-Layer Perceptron \\
$concat(\cdots)$, $\mathcal{C}(\cdots)$ & concatenation operator \\
$Transf(\cdot)$, $Tf(\cdot)$ & Transformer layer operator \\
$\mathbb{R}$ & real number  \\
$\odot$ &  Hadamard product \\
$\oplus$ &  element-wise sum \\


\hline
\hline
\tabincell{c}{Abbreviated \\ Terms} & Descriptions \\
\hline
CNN &  Convolutional Neural Network \\
TCN & Temporal Convolutional Neural Network \\
GNN & Graph Neural Network \\
{GCN} & {Graph Convolutional Network} \\
{BERT} & {Bidirectional Encoder Representations from Transformers} \\

$Norm(\cdot)$ & normalization \\
LN, ${LN}(\cdot)$ & Layer Normalization \\
BN, ${BN}(\cdot)$ & Batch Normalization \\
ReLU, ${ReLU}(\cdot)$ & Rectified Linear Unit \\
GELU, ${GELU}(\cdot)$ & Gaussian Error Linear Unit \\
VLP & Visual-Linguistic (Vision-Language) Pretraining \\
VidL & Video-and-Language \\
RoI & region-of-interest  \\
\hline
\end{tabular}
}
\end{center}

\end{table}

%% file: tables/tokens.tex
\begin{table}[!h]
\caption{Special/customized tokens. 
}
\label{table:tokens}
\begin{center}
\begin{tabular}{ l | l | l}
\hline
Tokens & Definitions &  References \\
\hline
\hline
\texttt{[CLS]} &  class &  \cite{shang2019pre, guo2020graphcodebert, zhu2020actbert, zhou2020unified}  \\
\texttt{[SEP]} &  separate &  \cite{shang2019pre, guo2020graphcodebert, zhu2020actbert, zhou2020unified}  \\
\texttt{[STOP]} & stop  &   \cite{zhou2020unified} \\
\texttt{[TASK]} & task  &  \cite{lu202012}  \\
\texttt{[MASK]} &  mask &  \cite{shang2019pre, su2019vl, guo2020graphcodebert, lin2020interbert}  \\

\texttt{[ACT]} &  action & \cite{zhu2020actbert}   \\

\texttt{[REGION]} & region  & \cite{zhu2020actbert}   \\

\texttt{[SOS]} & similar to \texttt{[SEP]}  &   \cite{rahman2021tribert} \\
\texttt{[>]} &  similar to \texttt{[SEP]} &  \cite{sun2019videobert}  \\
\texttt{[BOS]} & begin of sentence  & \cite{hu2021scaling}   \\
\texttt{[EOS]} & end of sentence  &  \cite{vaswani2017attention, hu2021scaling}  \\
\texttt{[PAD]} & padding  & \cite{vaswani2017attention}   \\
 \texttt{[END]} & end  &  \cite{su2019vl}   \\ 
\texttt{[TAB]} & table  &  \cite{chen2020open}  \\
\texttt{[TITLE]} &  title  &  \cite{chen2020open} \\
\texttt{[ROW]} & row  &  \cite{chen2020open}  \\
\texttt{[MAX]} &  maximum & \cite{chen2020open}   \\
\texttt{[MIN]} &  minimum &  \cite{chen2020open}  \\

\texttt{[IMG]}    & image   &      \cite{su2019vl, lu2019vilbert} \\

\texttt{[END]}    &  end  &      \cite{su2019vl} \\

\texttt{[FSN]} & fusion bottleneck  & \cite{nagrani2021attention}    \\

\hline
\end{tabular}
\end{center}

\end{table}

%% file: main.bbl
\begin{thebibliography}{100}
\providecommand{\url}[1]{#1}
\csname url@samestyle\endcsname
\providecommand{\newblock}{\relax}
\providecommand{\bibinfo}[2]{#2}
\providecommand{\BIBentrySTDinterwordspacing}{\spaceskip=0pt\relax}
\providecommand{\BIBentryALTinterwordstretchfactor}{4}
\providecommand{\BIBentryALTinterwordspacing}{\spaceskip=\fontdimen2\font plus
\BIBentryALTinterwordstretchfactor\fontdimen3\font minus
  \fontdimen4\font\relax}
\providecommand{\BIBforeignlanguage}[2]{{%
\expandafter\ifx\csname l@#1\endcsname\relax
\typeout{** WARNING: IEEEtran.bst: No hyphenation pattern has been}%
\typeout{** loaded for the language `#1'. Using the pattern for}%
\typeout{** the default language instead.}%
\else
\language=\csname l@#1\endcsname
\fi
#2}}
\providecommand{\BIBdecl}{\relax}
\BIBdecl

\bibitem{baltruvsaitis2018multimodal}
T.~Baltru{\v{s}}aitis, C.~Ahuja, and L.-P. Morency, ``Multimodal machine
  learning: A survey and taxonomy,'' \emph{TPAMI}, 2018.

\bibitem{vaswani2017attention}
A.~Vaswani, N.~Shazeer, N.~Parmar, J.~Uszkoreit, L.~Jones, A.~N. Gomez,
  {\L}.~Kaiser, and I.~Polosukhin, ``Attention is all you need,'' in
  \emph{NeurIPS}, 2017.

\bibitem{jaegle2021perceiver}
A.~Jaegle, F.~Gimeno, A.~Brock, A.~Zisserman, O.~Vinyals, and J.~Carreira,
  ``Perceiver: General perception with iterative attention,'' in \emph{ICML},
  2021.

\bibitem{devlin2018bert}
J.~Devlin, M.-W. Chang, K.~Lee, and K.~Toutanova, ``Bert: Pre-training of deep
  bidirectional transformers for language understanding,'' \emph{arXiv}, 2018.

\bibitem{dosovitskiy2020image}
A.~Dosovitskiy, L.~Beyer, A.~Kolesnikov, D.~Weissenborn, X.~Zhai,
  T.~Unterthiner, M.~Dehghani, M.~Minderer, G.~Heigold, S.~Gelly \emph{et~al.},
  ``An image is worth 16x16 words: Transformers for image recognition at
  scale,'' \emph{arXiv}, 2020.

\bibitem{carion2020end}
N.~Carion, F.~Massa, G.~Synnaeve, N.~Usunier, A.~Kirillov, and S.~Zagoruyko,
  ``End-to-end object detection with transformers,'' in \emph{ECCV}, 2020.

\bibitem{sun2019videobert}
C.~Sun, A.~Myers, C.~Vondrick, K.~Murphy, and C.~Schmid, ``Videobert: A joint
  model for video and language representation learning,'' in \emph{ICCV}, 2019.

\bibitem{chen2021speech}
J.~Chen, X.~Tan, Y.~Leng, J.~Xu, G.~Wen, T.~Qin, and T.-Y. Liu, ``Speech-t:
  Transducer for text to speech and beyond,'' \emph{NeurIPS}, 2021.

\bibitem{radford2021learning}
A.~Radford, J.~W. Kim, C.~Hallacy, A.~Ramesh, G.~Goh, S.~Agarwal, G.~Sastry,
  A.~Askell, P.~Mishkin, J.~Clark \emph{et~al.}, ``Learning transferable visual
  models from natural language supervision,'' \emph{arXiv}, 2021.

\bibitem{li2022clip}
M.~Li, R.~Xu, S.~Wang, L.~Zhou, X.~Lin, C.~Zhu, M.~Zeng, H.~Ji, and S.-F.
  Chang, ``Clip-event: Connecting text and images with event structures,''
  \emph{arXiv}, 2022.

\bibitem{zhang2020multimodal}
C.~Zhang, Z.~Yang, X.~He, and L.~Deng, ``Multimodal intelligence:
  Representation learning, information fusion, and applications,''
  \emph{JSTSP}, 2020.

\bibitem{rahate2022multimodal}
A.~Rahate, R.~Walambe, S.~Ramanna, and K.~Kotecha, ``Multimodal co-learning:
  Challenges, applications with datasets, recent advances and future
  directions,'' \emph{Information Fusion}, 2022.

\bibitem{hastie2009elements}
T.~Hastie, R.~Tibshirani, J.~H. Friedman, and J.~H. Friedman, \emph{The
  elements of statistical learning: data mining, inference, and
  prediction}.\hskip 1em plus 0.5em minus 0.4em\relax Springer, 2009, vol.~2.

\bibitem{parida2022beyond}
K.~K. Parida, S.~Srivastava, and G.~Sharma, ``Beyond mono to binaural:
  Generating binaural audio from mono audio with depth and cross modal
  attention,'' in \emph{WACV}, 2022.

\bibitem{qingyun2021cross}
F.~Qingyun, H.~Dapeng, and W.~Zhaokui, ``Cross-modality fusion transformer for
  multispectral object detection,'' \emph{arXiv}, 2021.

\bibitem{baevski2020wav2vec}
A.~Baevski, Y.~Zhou, A.~Mohamed, and M.~Auli, ``wav2vec 2.0: A framework for
  self-supervised learning of speech representations,'' \emph{NeurIPS}, 2020.

\bibitem{nagrani2020speech2action}
A.~Nagrani, C.~Sun, D.~Ross, R.~Sukthankar, C.~Schmid, and A.~Zisserman,
  ``Speech2action: Cross-modal supervision for action recognition,'' in
  \emph{CVPR}, 2020.

\bibitem{chen2020open}
W.~Chen, M.-W. Chang, E.~Schlinger, W.~Wang, and W.~W. Cohen, ``Open question
  answering over tables and text,'' \emph{arXiv}, 2020.

\bibitem{guo2021general}
Y.~Guo, L.~Gao, X.~Wang, Y.~Hu, X.~Xu, X.~Lu, H.~T. Shen, and J.~Song, ``From
  general to specific: Informative scene graph generation via balance
  adjustment,'' in \emph{ICCV}, 2021.

\bibitem{gupta2020layouttransformer}
K.~Gupta, J.~Lazarow, A.~Achille, L.~Davis, V.~Mahadevan, and A.~Shrivastava,
  ``Layouttransformer: Layout generation and completion with self-attention,''
  \emph{arXiv}, 2020.

\bibitem{yang2021layouttransformer}
C.-F. Yang, W.-C. Fan, F.-E. Yang, and Y.-C.~F. Wang, ``Layouttransformer:
  Scene layout generation with conceptual and spatial diversity,'' in
  \emph{CVPR}, 2021.

\bibitem{Li_2022_CVPR_SGTR}
R.~Li, S.~Zhang, and X.~He, ``Sgtr: End-to-end scene graph generation with
  transformer,'' in \emph{CVPR}, 2022.

\bibitem{esser2021taming}
P.~Esser, R.~Rombach, and B.~Ommer, ``Taming transformers for high-resolution
  image synthesis,'' in \emph{CVPR}, 2021.

\bibitem{cai2021sadga}
R.~Cai, J.~Yuan, B.~Xu, and Z.~Hao, ``Sadga: Structure-aware dual graph
  aggregation network for text-to-sql,'' \emph{NeurIPS}, 2021.

\bibitem{song2022speech}
Y.~Song, R.~C.-W. Wong, X.~Zhao, and D.~Jiang, ``Speech-to-sql: Towards
  speech-driven sql query generation from natural language question,''
  \emph{arXiv}, 2022.

\bibitem{salvador2021revamping}
A.~Salvador, E.~Gundogdu, L.~Bazzani, and M.~Donoser, ``Revamping cross-modal
  recipe retrieval with hierarchical transformers and self-supervised
  learning,'' in \emph{CVPR}, 2021.

\bibitem{zhao2021proto}
Z.~Zhao, K.~Samel, B.~Chen \emph{et~al.}, ``Proto: Program-guided transformer
  for program-guided tasks,'' in \emph{NeurIPS}, 2021.

\bibitem{zhou2021improving}
H.~Zhou, W.~Zhou, W.~Qi, J.~Pu, and H.~Li, ``Improving sign language
  translation with monolingual data by sign back-translation,'' in \emph{CVPR},
  2021.

\bibitem{varol2021read}
G.~Varol, L.~Momeni, S.~Albanie, T.~Afouras, and A.~Zisserman, ``Read and
  attend: Temporal localisation in sign language videos,'' in \emph{CVPR},
  2021.

\bibitem{bull2021aligning}
H.~Bull, T.~Afouras, G.~Varol, S.~Albanie, L.~Momeni, and A.~Zisserman,
  ``Aligning subtitles in sign language videos,'' \emph{arXiv}, 2021.

\bibitem{zhao20213dvg}
L.~Zhao, D.~Cai, L.~Sheng, and D.~Xu, ``3dvg-transformer: Relation modeling for
  visual grounding on point clouds,'' in \emph{ICCV}, 2021.

\bibitem{marino2021krisp}
K.~Marino, X.~Chen, D.~Parikh, A.~Gupta, and M.~Rohrbach, ``Krisp: Integrating
  implicit and symbolic knowledge for open-domain knowledge-based vqa,'' in
  \emph{CVPR}, 2021.

\bibitem{ammanabrolu2021learning}
P.~Ammanabrolu and M.~O. Riedl, ``Learning knowledge graph-based world models
  of textual environments,'' \emph{arXiv}, 2021.

\bibitem{zhu2022multi}
X.~Zhu, Z.~Li, X.~Wang, X.~Jiang, P.~Sun, X.~Wang, Y.~Xiao, and N.~J. Yuan,
  ``Multi-modal knowledge graph construction and application: A survey,''
  \emph{arXiv}, 2022.

\bibitem{xu2018sketchmate}
P.~Xu, Y.~Huang, T.~Yuan, K.~Pang, Y.-Z. Song, T.~Xiang, T.~M. Hospedales,
  Z.~Ma, and J.~Guo, ``Sketchmate: Deep hashing for million-scale human sketch
  retrieval,'' in \emph{CVPR}, 2018.

\bibitem{xu2020deep}
P.~Xu, Z.~Song, Q.~Yin, Y.-Z. Song, and L.~Wang, ``Deep self-supervised
  representation learning for free-hand sketch,'' \emph{TCSVT}, 2020.

\bibitem{xu2020fine}
P.~Xu, K.~Liu, T.~Xiang, T.~M. Hospedales, Z.~Ma, J.~Guo, and Y.-Z. Song,
  ``Fine-grained instance-level sketch-based video retrieval,'' \emph{TCSVT},
  2020.

\bibitem{vinker2022clipasso}
Y.~Vinker, E.~Pajouheshgar, J.~Y. Bo, R.~C. Bachmann, A.~H. Bermano,
  D.~Cohen-Or, A.~Zamir, and A.~Shamir, ``Clipasso: Semantically-aware object
  sketching,'' \emph{arXiv}, 2022.

\bibitem{fan2021faceformer}
Y.~Fan, Z.~Lin, J.~Saito, W.~Wang, and T.~Komura, ``Faceformer: Speech-driven
  3d facial animation with transformers,'' \emph{arXiv}, 2021.

\bibitem{shin20193d}
D.~Shin, Z.~Ren, E.~B. Sudderth, and C.~C. Fowlkes, ``3d scene reconstruction
  with multi-layer depth and epipolar transformers,'' in \emph{ICCV}, 2019.

\bibitem{lin2021end}
K.~Lin, L.~Wang, and Z.~Liu, ``End-to-end human pose and mesh reconstruction
  with transformers,'' in \emph{CVPR}, 2021.

\bibitem{xu2020layoutlmv2}
Y.~Xu, Y.~Xu, T.~Lv, L.~Cui, F.~Wei, G.~Wang, Y.~Lu, D.~Florencio, C.~Zhang,
  W.~Che \emph{et~al.}, ``Layoutlmv2: Multi-modal pre-training for
  visually-rich document understanding,'' \emph{arXiv}, 2020.

\bibitem{beltagy2020longformer}
I.~Beltagy, M.~E. Peters, and A.~Cohan, ``Longformer: The long-document
  transformer,'' \emph{arXiv}, 2020.

\bibitem{guo2020graphcodebert}
D.~Guo, S.~Ren, S.~Lu, Z.~Feng, D.~Tang, S.~Liu, L.~Zhou, N.~Duan,
  A.~Svyatkovskiy, S.~Fu \emph{et~al.}, ``Graphcodebert: Pre-training code
  representations with data flow,'' \emph{arXiv}, 2020.

\bibitem{zugner2021language}
D.~Z{\"u}gner, T.~Kirschstein, M.~Catasta, J.~Leskovec, and S.~G{\"u}nnemann,
  ``Language-agnostic representation learning of source code from structure and
  context,'' \emph{arXiv}, 2021.

\bibitem{gavrilyuk2020actor}
K.~Gavrilyuk, R.~Sanford, M.~Javan, and C.~G. Snoek, ``Actor-transformers for
  group activity recognition,'' in \emph{CVPR}, 2020.

\bibitem{shang2019pre}
J.~Shang, T.~Ma, C.~Xiao, and J.~Sun, ``Pre-training of graph augmented
  transformers for medication recommendation,'' \emph{arXiv}, 2019.

\bibitem{lin2021survey}
T.~Lin, Y.~Wang, X.~Liu, and X.~Qiu, ``A survey of transformers,''
  \emph{arXiv}, 2021.

\bibitem{tay2020efficient}
Y.~Tay, M.~Dehghani, D.~Bahri, and D.~Metzler, ``Efficient transformers: A
  survey,'' \emph{arXiv}, 2020.

\bibitem{bracsoveanu2020visualizing}
A.~M. Bra{\c{s}}oveanu and R.~Andonie, ``Visualizing transformers for nlp: a
  brief survey,'' in \emph{International Conference Information Visualisation
  (IV)}, 2020.

\bibitem{khan2021transformers}
S.~Khan, M.~Naseer, M.~Hayat, S.~W. Zamir, F.~S. Khan, and M.~Shah,
  ``Transformers in vision: A survey,'' \emph{arXiv}, 2021.

\bibitem{liu2021survey}
Y.~Liu, Y.~Zhang, Y.~Wang, F.~Hou, J.~Yuan, J.~Tian, Y.~Zhang, Z.~Shi, J.~Fan,
  and Z.~He, ``A survey of visual transformers,'' \emph{arXiv}, 2021.

\bibitem{han2020survey}
K.~Han, Y.~Wang, H.~Chen, X.~Chen, J.~Guo, Z.~Liu, Y.~Tang, A.~Xiao, C.~Xu,
  Y.~Xu \emph{et~al.}, ``A survey on vision transformer,'' \emph{arXiv}, 2020.

\bibitem{xu2022transformers}
Y.~Xu, H.~Wei, M.~Lin, Y.~Deng, K.~Sheng, M.~Zhang, F.~Tang, W.~Dong, F.~Huang,
  and C.~Xu, ``Transformers in computational visual media: A survey,''
  \emph{Computational Visual Media}, 2022.

\bibitem{shamshad2022transformers}
F.~Shamshad, S.~Khan, S.~W. Zamir, M.~H. Khan, M.~Hayat, F.~S. Khan, and H.~Fu,
  ``Transformers in medical imaging: A survey,'' \emph{arXiv}, 2022.

\bibitem{selva2022video}
J.~Selva, A.~S. Johansen, S.~Escalera, K.~Nasrollahi, T.~B. Moeslund, and
  A.~Clap{\'e}s, ``Video transformers: A survey,'' \emph{arXiv}, 2022.

\bibitem{ruan2021survey}
L.~Ruan and Q.~Jin, ``Survey: Transformer based video-language pre-training,''
  \emph{arXiv}, 2021.

\bibitem{chen2022vlp}
F.~Chen, D.~Zhang, M.~Han, X.~Chen, J.~Shi, S.~Xu, and B.~Xu, ``Vlp: A survey
  on vision-language pre-training,'' \emph{arXiv}, 2022.

\bibitem{li2022vision}
F.~Li, H.~Zhang, Y.-F. Zhang, S.~Liu, J.~Guo, L.~M. Ni, P.~Zhang, and L.~Zhang,
  ``Vision-language intelligence: Tasks, representation learning, and large
  models,'' \emph{arXiv}, 2022.

\bibitem{wu1999multimodal}
L.~Wu, S.~L. Oviatt, and P.~R. Cohen, ``Multimodal integration-a statistical
  view,'' \emph{TMM}, 1999.

\bibitem{guo2019deep}
W.~Guo, J.~Wang, and S.~Wang, ``Deep multimodal representation learning: A
  survey,'' \emph{IEEE Access}, 2019.

\bibitem{yuhas1989integration}
B.~P. Yuhas, M.~H. Goldstein, and T.~J. Sejnowski, ``Integration of acoustic
  and visual speech signals using neural networks,'' \emph{IEEE Communications
  Magazine}, 1989.

\bibitem{lazarus1976multimodal}
A.~A. Lazarus \emph{et~al.}, \emph{Multimodal behavior therapy}.\hskip 1em plus
  0.5em minus 0.4em\relax Springer, 1976.

\bibitem{feng2020deep}
D.~Feng, C.~Haase-Sch{\"u}tz, L.~Rosenbaum, H.~Hertlein, C.~Glaeser, F.~Timm,
  W.~Wiesbeck, and K.~Dietmayer, ``Deep multi-modal object detection and
  semantic segmentation for autonomous driving: Datasets, methods, and
  challenges,'' \emph{TITS}, 2020.

\bibitem{liu2021multimodal}
Y.~Liu, J.~Zhang, L.~Fang, Q.~Jiang, and B.~Zhou, ``Multimodal motion
  prediction with stacked transformers,'' in \emph{CVPR}, 2021.

\bibitem{moudgil2021soat}
A.~Moudgil, A.~Majumdar, H.~Agrawal, S.~Lee, and D.~Batra, ``Soat: A scene-and
  object-aware transformer for vision-and-language navigation,''
  \emph{NeurIPS}, 2021.

\bibitem{lv2021progressive}
F.~Lv, X.~Chen, Y.~Huang, L.~Duan, and G.~Lin, ``Progressive modality
  reinforcement for human multimodal emotion recognition from unaligned
  multimodal sequences,'' in \emph{CVPR}, 2021.

\bibitem{zellers2021merlot}
R.~Zellers, X.~Lu, J.~Hessel, Y.~Yu, J.~S. Park, J.~Cao, A.~Farhadi, and
  Y.~Choi, ``Merlot: Multimodal neural script knowledge models,'' \emph{arXiv},
  2021.

\bibitem{hasan2021humor}
M.~K. Hasan, S.~Lee, W.~Rahman, A.~Zadeh, R.~Mihalcea, L.-P. Morency, and
  E.~Hoque, ``Humor knowledge enriched transformer for understanding multimodal
  humor,'' in \emph{AAAI}, 2021.

\bibitem{brown2021face}
A.~Brown, V.~Kalogeiton, and A.~Zisserman, ``Face, body, voice: Video
  person-clustering with multiple modalities,'' \emph{arXiv}, 2021.

\bibitem{yu2022commercemm}
L.~Yu, J.~Chen, A.~Sinha, M.~M. Wang, H.~Chen, T.~L. Berg, and N.~Zhang,
  ``Commercemm: Large-scale commerce multimodal representation learning with
  omni retrieval,'' \emph{arXiv}, 2022.

\bibitem{chen2021topological}
K.~Chen, J.~K. Chen, J.~Chuang, M.~V{\'a}zquez, and S.~Savarese, ``Topological
  planning with transformers for vision-and-language navigation,'' in
  \emph{CVPR}, 2021.

\bibitem{hong2021vln}
Y.~Hong, Q.~Wu, Y.~Qi, C.~Rodriguez-Opazo, and S.~Gould, ``Vln bert: A
  recurrent vision-and-language bert for navigation,'' in \emph{CVPR}, 2021.

\bibitem{zhang2021curriculum}
J.~Zhang, J.~Fan, J.~Peng \emph{et~al.}, ``Curriculum learning for
  vision-and-language navigation,'' in \emph{NeurIPS}, 2021.

\bibitem{qi2021road}
Y.~Qi, Z.~Pan, Y.~Hong, M.-H. Yang, A.~van~den Hengel, and Q.~Wu, ``The road to
  know-where: An object-and-room informed sequential bert for indoor
  vision-language navigation,'' in \emph{ICCV}, 2021.

\bibitem{chen2021semantic}
C.~Chen, Z.~Al-Halah, and K.~Grauman, ``Semantic audio-visual navigation,'' in
  \emph{CVPR}, 2021.

\bibitem{ren2021learning}
S.~Ren, Y.~Du, J.~Lv, G.~Han, and S.~He, ``Learning from the master: Distilling
  cross-modal advanced knowledge for lip reading,'' in \emph{CVPR}, 2021.

\bibitem{xu2022deep}
P.~Xu, T.~M. Hospedales, Q.~Yin, Y.-Z. Song, T.~Xiang, and L.~Wang, ``Deep
  learning for free-hand sketch: A survey,'' \emph{TPAMI}, 2022.

\bibitem{li2020behrt}
Y.~Li, S.~Rao, J.~R.~A. Solares, A.~Hassaine, R.~Ramakrishnan, D.~Canoy,
  Y.~Zhu, K.~Rahimi, and G.~Salimi-Khorshidi, ``Behrt: transformer for
  electronic health records,'' \emph{Scientific reports}, 2020.

\bibitem{li2020comparison}
Y.~Li, H.~Wang, and Y.~Luo, ``A comparison of pre-trained vision-and-language
  models for multimodal representation learning across medical images and
  reports,'' in \emph{BIBM}, 2020.

\bibitem{xu2021deepchange}
P.~Xu and X.~Zhu, ``Deepchange: A large long-term person re-identification
  benchmark with clothes change,'' \emph{arXiv}, 2021.

\bibitem{tsimpoukelli2021multimodal}
M.~Tsimpoukelli, J.~Menick, S.~Cabi, S.~Eslami, O.~Vinyals, and F.~Hill,
  ``Multimodal few-shot learning with frozen language models,'' \emph{NeurIPS},
  2021.

\bibitem{sung2022vl}
Y.-L. Sung, J.~Cho, and M.~Bansal, ``Vl-adapter: Parameter-efficient transfer
  learning for vision-and-language tasks,'' in \emph{CVPR}, 2022.

\bibitem{alayrac2022flamingo}
J.-B. Alayrac, J.~Donahue, P.~Luc, A.~Miech, I.~Barr, Y.~Hasson, K.~Lenc,
  A.~Mensch, K.~Millican, M.~Reynolds \emph{et~al.}, ``Flamingo: a visual
  language model for few-shot learning,'' \emph{NeurIPS}, 2022.

\bibitem{wang2022image}
W.~Wang, H.~Bao, L.~Dong, J.~Bjorck, Z.~Peng, Q.~Liu, K.~Aggarwal, O.~K.
  Mohammed, S.~Singhal, S.~Som \emph{et~al.}, ``Image as a foreign language:
  Beit pretraining for all vision and vision-language tasks,'' \emph{arXiv},
  2022.

\bibitem{chen2022pali}
X.~Chen, X.~Wang, S.~Changpinyo, A.~Piergiovanni, P.~Padlewski, D.~Salz,
  S.~Goodman, A.~Grycner, B.~Mustafa, L.~Beyer \emph{et~al.}, ``Pali: A
  jointly-scaled multilingual language-image model,'' \emph{arXiv}, 2022.

\bibitem{lewis2019bart}
M.~Lewis, Y.~Liu, N.~Goyal, M.~Ghazvininejad, A.~Mohamed, O.~Levy, V.~Stoyanov,
  and L.~Zettlemoyer, ``Bart: Denoising sequence-to-sequence pre-training for
  natural language generation, translation, and comprehension,'' \emph{arXiv},
  2019.

\bibitem{radford2018improving}
A.~Radford, K.~Narasimhan, T.~Salimans, and I.~Sutskever, ``Improving language
  understanding by generative pre-training,'' 2018.

\bibitem{dai2019transformer}
Z.~Dai, Z.~Yang, Y.~Yang, J.~Carbonell, Q.~V. Le, and R.~Salakhutdinov,
  ``Transformer-xl: Attentive language models beyond a fixed-length context,''
  \emph{arXiv}, 2019.

\bibitem{yang2019xlnet}
Z.~Yang, Z.~Dai, Y.~Yang, J.~Carbonell, R.~R. Salakhutdinov, and Q.~V. Le,
  ``Xlnet: Generalized autoregressive pretraining for language understanding,''
  \emph{NeurIPS}, 2019.

\bibitem{chen2020generative}
M.~Chen, A.~Radford, R.~Child, J.~Wu, H.~Jun, D.~Luan, and I.~Sutskever,
  ``Generative pretraining from pixels,'' in \emph{ICML}, 2020.

\bibitem{chen2021pre}
H.~Chen, Y.~Wang, T.~Guo, C.~Xu, Y.~Deng, Z.~Liu, S.~Ma, C.~Xu, C.~Xu, and
  W.~Gao, ``Pre-trained image processing transformer,'' in \emph{CVPR}, 2021.

\bibitem{touvron2021training}
H.~Touvron, M.~Cord, M.~Douze, F.~Massa, A.~Sablayrolles, and H.~J{\'e}gou,
  ``Training data-efficient image transformers \& distillation through
  attention,'' in \emph{ICML}, 2021.

\bibitem{beal2020toward}
J.~Beal, E.~Kim, E.~Tzeng, D.~H. Park, A.~Zhai, and D.~Kislyuk, ``Toward
  transformer-based object detection,'' \emph{arXiv}, 2020.

\bibitem{liu2021swin}
Z.~Liu, Y.~Lin, Y.~Cao, H.~Hu, Y.~Wei, Z.~Zhang, S.~Lin, and B.~Guo, ``Swin
  transformer: Hierarchical vision transformer using shifted windows,''
  \emph{arXiv}, 2021.

\bibitem{chen2021empirical}
X.~Chen, S.~Xie, and K.~He, ``An empirical study of training self-supervised
  vision transformers,'' \emph{arXiv}, 2021.

\bibitem{caron2021emerging}
M.~Caron, H.~Touvron, I.~Misra, H.~J{\'e}gou, J.~Mairal, P.~Bojanowski, and
  A.~Joulin, ``Emerging properties in self-supervised vision transformers,''
  \emph{arXiv}, 2021.

\bibitem{bao2021beit}
H.~Bao, L.~Dong, and F.~Wei, ``Beit: Bert pre-training of image transformers,''
  \emph{arXiv}, 2021.

\bibitem{paul2021vision}
S.~Paul and P.-Y. Chen, ``Vision transformers are robust learners,''
  \emph{arXiv}, 2021.

\bibitem{raghu2021vision}
M.~Raghu, T.~Unterthiner, S.~Kornblith, C.~Zhang, and A.~Dosovitskiy, ``Do
  vision transformers see like convolutional neural networks?'' \emph{NeurIPS},
  2021.

\bibitem{cao2022understand}
S.~Cao, P.~Xu, and D.~A. Clifton, ``How to understand masked autoencoders,''
  \emph{arXiv}, 2022.

\bibitem{lu2019vilbert}
J.~Lu, D.~Batra, D.~Parikh, and S.~Lee, ``Vilbert: Pretraining task-agnostic
  visiolinguistic representations for vision-and-language tasks,''
  \emph{arXiv}, 2019.

\bibitem{tan2019lxmert}
H.~Tan and M.~Bansal, ``Lxmert: Learning cross-modality encoder representations
  from transformers,'' \emph{arXiv}, 2019.

\bibitem{li2019visualbert}
L.~H. Li, M.~Yatskar, D.~Yin, C.-J. Hsieh, and K.-W. Chang, ``Visualbert: A
  simple and performant baseline for vision and language,'' \emph{arXiv}, 2019.

\bibitem{su2019vl}
W.~Su, X.~Zhu, Y.~Cao, B.~Li, L.~Lu, F.~Wei, and J.~Dai, ``Vl-bert:
  Pre-training of generic visual-linguistic representations,'' \emph{arXiv},
  2019.

\bibitem{chen2020uniter}
Y.-C. Chen, L.~Li, L.~Yu, A.~El~Kholy, F.~Ahmed, Z.~Gan, Y.~Cheng, and J.~Liu,
  ``Uniter: Universal image-text representation learning,'' in \emph{ECCV},
  2020.

\bibitem{sun2019learning}
C.~Sun, F.~Baradel, K.~Murphy, and C.~Schmid, ``Learning video representations
  using contrastive bidirectional transformer,'' \emph{arXiv}, 2019.

\bibitem{li2020unicoder}
G.~Li, N.~Duan, Y.~Fang, M.~Gong, and D.~Jiang, ``Unicoder-vl: A universal
  encoder for vision and language by cross-modal pre-training,'' in
  \emph{AAAI}, 2020.

\bibitem{alberti2019fusion}
C.~Alberti, J.~Ling, M.~Collins, and D.~Reitter, ``Fusion of detected objects
  in text for visual question answering,'' \emph{arXiv}, 2019.

\bibitem{zhou2020unified}
L.~Zhou, H.~Palangi, L.~Zhang, H.~Hu, J.~Corso, and J.~Gao, ``Unified
  vision-language pre-training for image captioning and vqa,'' in \emph{AAAI},
  2020.

\bibitem{lu202012}
J.~Lu, V.~Goswami, M.~Rohrbach, D.~Parikh, and S.~Lee, ``12-in-1: Multi-task
  vision and language representation learning,'' in \emph{CVPR}, 2020.

\bibitem{li2020oscar}
X.~Li, X.~Yin, C.~Li, P.~Zhang, X.~Hu, L.~Zhang, L.~Wang, H.~Hu, L.~Dong,
  F.~Wei \emph{et~al.}, ``Oscar: Object-semantics aligned pre-training for
  vision-language tasks,'' in \emph{ECCV}, 2020.

\bibitem{huang2020pixel}
Z.~Huang, Z.~Zeng, B.~Liu, D.~Fu, and J.~Fu, ``Pixel-bert: Aligning image
  pixels with text by deep multi-modal transformers,'' \emph{arXiv}, 2020.

\bibitem{zhu2020actbert}
L.~Zhu and Y.~Yang, ``Actbert: Learning global-local video-text
  representations,'' in \emph{CVPR}, 2020.

\bibitem{qi2020imagebert}
D.~Qi, L.~Su, J.~Song, E.~Cui, T.~Bharti, and A.~Sacheti, ``Imagebert:
  Cross-modal pre-training with large-scale weak-supervised image-text data,''
  \emph{arXiv}, 2020.

\bibitem{li2020hero}
L.~Li, Y.-C. Chen, Y.~Cheng, Z.~Gan, L.~Yu, and J.~Liu, ``Hero: Hierarchical
  encoder for video+ language omni-representation pre-training,'' \emph{arXiv},
  2020.

\bibitem{luo2020univl}
H.~Luo, L.~Ji, B.~Shi, H.~Huang, N.~Duan, T.~Li, J.~Li, T.~Bharti, and M.~Zhou,
  ``Univl: A unified video and language pre-training model for multimodal
  understanding and generation,'' \emph{arXiv}, 2020.

\bibitem{xu2021simple}
M.~Xu, Z.~Zhang, F.~Wei, Y.~Lin, Y.~Cao, H.~Hu, and X.~Bai, ``A simple baseline
  for zero-shot semantic segmentation with pre-trained vision-language model,''
  \emph{arXiv}, 2021.

\bibitem{jia2021scaling}
C.~Jia, Y.~Yang, Y.~Xia, Y.-T. Chen, Z.~Parekh, H.~Pham, Q.~V. Le, Y.~Sung,
  Z.~Li, and T.~Duerig, ``Scaling up visual and vision-language representation
  learning with noisy text supervision,'' \emph{arXiv}, 2021.

\bibitem{wang2022clip}
Z.~Wang, N.~Codella, Y.-C. Chen, L.~Zhou, J.~Yang, X.~Dai, B.~Xiao, H.~You,
  S.-F. Chang, and L.~Yuan, ``Clip-td: Clip targeted distillation for
  vision-language tasks,'' \emph{arXiv}, 2022.

\bibitem{li2021alignbefore}
J.~Li, R.~Selvaraju, A.~Gotmare, S.~Joty, C.~Xiong, and S.~C.~H. Hoi, ``Align
  before fuse: Vision and language representation learning with momentum
  distillation,'' \emph{NeurIPS}, 2021.

\bibitem{yu2022coca}
J.~Yu, Z.~Wang, V.~Vasudevan, L.~Yeung, M.~Seyedhosseini, and Y.~Wu, ``Coca:
  Contrastive captioners are image-text foundation models,'' \emph{arXiv},
  2022.

\bibitem{sharma2018conceptual}
P.~Sharma, N.~Ding, S.~Goodman, and R.~Soricut, ``Conceptual captions: A
  cleaned, hypernymed, image alt-text dataset for automatic image captioning,''
  in \emph{ACL}, 2018.

\bibitem{lin2014microsoft}
T.-Y. Lin, M.~Maire, S.~Belongie, J.~Hays, P.~Perona, D.~Ramanan,
  P.~Doll{\'a}r, and C.~L. Zitnick, ``Microsoft coco: Common objects in
  context,'' in \emph{ECCV}, 2014.

\bibitem{antol2015vqa}
S.~Antol, A.~Agrawal, J.~Lu, M.~Mitchell, D.~Batra, C.~L. Zitnick, and
  D.~Parikh, ``Vqa: Visual question answering,'' in \emph{ICCV}, 2015.

\bibitem{krishna2017visual}
R.~Krishna, Y.~Zhu, O.~Groth, J.~Johnson, K.~Hata, J.~Kravitz, S.~Chen,
  Y.~Kalantidis, L.-J. Li, D.~A. Shamma \emph{et~al.}, ``Visual genome:
  Connecting language and vision using crowdsourced dense image annotations,''
  \emph{IJCV}, 2017.

\bibitem{ordonez2011im2text}
V.~Ordonez, G.~Kulkarni, and T.~Berg, ``Im2text: Describing images using 1
  million captioned photographs,'' \emph{NeurIPS}, 2011.

\bibitem{kayser2021vil}
M.~Kayser, O.-M. Camburu, L.~Salewski, C.~Emde, V.~Do, Z.~Akata, and
  T.~Lukasiewicz, ``e-vil: A dataset and benchmark for natural language
  explanations in vision-language tasks,'' \emph{arXiv}, 2021.

\bibitem{gamper2021multiple}
J.~Gamper and N.~Rajpoot, ``Multiple instance captioning: Learning
  representations from histopathology textbooks and articles,'' in \emph{CVPR},
  2021.

\bibitem{li2021adversarial}
L.~Li, J.~Lei, Z.~Gan, and J.~Liu, ``Adversarial vqa: A new benchmark for
  evaluating the robustness of vqa models,'' \emph{arXiv}, 2021.

\bibitem{talmor2021multimodalqa}
A.~Talmor, O.~Yoran, A.~Catav, D.~Lahav, Y.~Wang, A.~Asai, G.~Ilharco,
  H.~Hajishirzi, and J.~Berant, ``Multimodalqa: Complex question answering over
  text, tables and images,'' \emph{arXiv}, 2021.

\bibitem{li2021value}
L.~Li, J.~Lei, Z.~Gan, L.~Yu, Y.-C. Chen, R.~Pillai, Y.~Cheng, L.~Zhou, X.~E.
  Wang, W.~Y. Wang \emph{et~al.}, ``Value: A multi-task benchmark for
  video-and-language understanding evaluation,'' \emph{arXiv}, 2021.

\bibitem{wu2021fashion}
H.~Wu, Y.~Gao, X.~Guo, Z.~Al-Halah, S.~Rennie, K.~Grauman, and R.~Feris,
  ``Fashion iq: A new dataset towards retrieving images by natural language
  feedback,'' in \emph{CVPR}, 2021.

\bibitem{afouras2018deep}
T.~Afouras, J.~S. Chung, A.~Senior, O.~Vinyals, and A.~Zisserman, ``Deep
  audio-visual speech recognition,'' \emph{TPAMI}, 2018.

\bibitem{krishna2017dense}
R.~Krishna, K.~Hata, F.~Ren, L.~Fei-Fei, and J.~Carlos~Niebles,
  ``Dense-captioning events in videos,'' in \emph{ICCV}, 2017.

\bibitem{das2017visual}
A.~Das, S.~Kottur, K.~Gupta, A.~Singh, D.~Yadav, J.~M. Moura, D.~Parikh, and
  D.~Batra, ``Visual dialog,'' in \emph{CVPR}, 2017.

\bibitem{zhan2021product1m}
X.~Zhan, Y.~Wu, X.~Dong, Y.~Wei, M.~Lu, Y.~Zhang, H.~Xu, and X.~Liang,
  ``Product1m: Towards weakly supervised instance-level product retrieval via
  cross-modal pretraining,'' in \emph{ICCV}, 2021.

\bibitem{changpinyo2021conceptual}
S.~Changpinyo, P.~Sharma, N.~Ding, and R.~Soricut, ``Conceptual 12m: Pushing
  web-scale image-text pre-training to recognize long-tail visual concepts,''
  in \emph{CVPR}, 2021.

\bibitem{huo2021wenlan}
Y.~Huo, M.~Zhang, G.~Liu, H.~Lu, Y.~Gao, G.~Yang, J.~Wen, H.~Zhang, B.~Xu,
  W.~Zheng \emph{et~al.}, ``Wenlan: Bridging vision and language by large-scale
  multi-modal pre-training,'' \emph{arXiv}, 2021.

\bibitem{yang2021just}
A.~Yang, A.~Miech, J.~Sivic, I.~Laptev, and C.~Schmid, ``Just ask: Learning to
  answer questions from millions of narrated videos,'' in \emph{ICCV}, 2021.

\bibitem{miech2019howto100m}
A.~Miech, D.~Zhukov, J.-B. Alayrac, M.~Tapaswi, I.~Laptev, and J.~Sivic,
  ``Howto100m: Learning a text-video embedding by watching hundred million
  narrated video clips,'' in \emph{ICCV}, 2019.

\bibitem{hu2021scaling}
X.~Hu, Z.~Gan, J.~Wang, Z.~Yang, Z.~Liu, Y.~Lu, and L.~Wang, ``Scaling up
  vision-language pre-training for image captioning,'' \emph{arXiv}, 2021.

\bibitem{schuhmann2021laion}
C.~Schuhmann, R.~Vencu, R.~Beaumont, R.~Kaczmarczyk, C.~Mullis, A.~Katta,
  T.~Coombes, J.~Jitsev, and A.~Komatsuzaki, ``Laion-400m: Open dataset of
  clip-filtered 400 million image-text pairs,'' \emph{arXiv}, 2021.

\bibitem{yun2021pano}
H.~Yun, Y.~Yu, W.~Yang, K.~Lee, and G.~Kim, ``Pano-avqa: Grounded audio-visual
  question answering on 360deg videos,'' in \emph{ICCV}, 2021.

\bibitem{morgado2020learning}
P.~Morgado, Y.~Li, and N.~Vasconcelos, ``Learning representations from
  audio-visual spatial alignment,'' \emph{arXiv}, 2020.

\bibitem{li2021ai}
R.~Li, S.~Yang, D.~A. Ross, and A.~Kanazawa, ``Ai choreographer: Music
  conditioned 3d dance generation with aist++,'' in \emph{ICCV}, 2021.

\bibitem{achlioptas2021artemis}
P.~Achlioptas, M.~Ovsjanikov, K.~Haydarov, M.~Elhoseiny, and L.~J. Guibas,
  ``Artemis: Affective language for visual art,'' in \emph{CVPR}, 2021.

\bibitem{liang2021multibench}
P.~P. Liang, Y.~Lyu, X.~Fan, Z.~Wu, Y.~Cheng, J.~Wu, L.~Y. Chen, P.~Wu, M.~A.
  Lee, Y.~Zhu \emph{et~al.}, ``Multibench: Multiscale benchmarks for multimodal
  representation learning,'' \emph{arXiv}, 2021.

\bibitem{liu2021image}
Z.~Liu, C.~Rodriguez-Opazo, D.~Teney, and S.~Gould, ``Image retrieval on
  real-life images with pre-trained vision-and-language models,'' in
  \emph{ICCV}, 2021.

\bibitem{guhur2021airbert}
P.-L. Guhur, M.~Tapaswi, S.~Chen, I.~Laptev, and C.~Schmid, ``Airbert:
  In-domain pretraining for vision-and-language navigation,'' in \emph{ICCV},
  2021.

\bibitem{sawhney2021multimodal}
R.~Sawhney, M.~Goyal, P.~Goel, P.~Mathur, and R.~Shah, ``Multimodal
  multi-speaker merger \& acquisition financial modeling: A new task, dataset,
  and neural baselines,'' in \emph{ACL-IJCNLP}, 2021.

\bibitem{zhang2021x}
J.~Zhang, M.~Zheng, M.~Boyd, and E.~Ohn-Bar, ``X-world: Accessibility, vision,
  and autonomy meet,'' in \emph{ICCV}, 2021.

\bibitem{zhang2021multimet}
D.~Zhang, M.~Zhang, H.~Zhang, L.~Yang, and H.~Lin, ``Multimet: A multimodal
  dataset for metaphor understanding,'' in \emph{ACL-IJCNLP}, 2021.

\bibitem{kiela2020hateful}
D.~Kiela, H.~Firooz, A.~Mohan, V.~Goswami, A.~Singh, P.~Ringshia, and
  D.~Testuggine, ``The hateful memes challenge: Detecting hate speech in
  multimodal memes,'' \emph{arXiv}, 2020.

\bibitem{zhou2018towards}
L.~Zhou, C.~Xu, and J.~J. Corso, ``Towards automatic learning of procedures
  from web instructional videos,'' in \emph{AAAI}, 2018.

\bibitem{malmaud2015s}
J.~Malmaud, J.~Huang, V.~Rathod, N.~Johnston, A.~Rabinovich, and K.~Murphy,
  ``What's cookin'? interpreting cooking videos using text, speech and
  vision,'' \emph{arXiv}, 2015.

\bibitem{bronstein2021geometric}
M.~M. Bronstein, J.~Bruna, T.~Cohen, and P.~Veli{\v{c}}kovi{\'c}, ``Geometric
  deep learning: Grids, groups, graphs, geodesics, and gauges,'' \emph{arXiv},
  2021.

\bibitem{dwivedi2020generalization}
V.~P. Dwivedi and X.~Bresson, ``A generalization of transformer networks to
  graphs,'' \emph{arXiv}, 2020.

\bibitem{he2016deep}
K.~He, X.~Zhang, S.~Ren, and J.~Sun, ``Deep residual learning for image
  recognition,'' in \emph{CVPR}, 2016.

\bibitem{ioffe2015batch}
S.~Ioffe and C.~Szegedy, ``Batch normalization: Accelerating deep network
  training by reducing internal covariate shift,'' in \emph{ICML}, 2015.

\bibitem{ba2016layer}
J.~L. Ba, J.~R. Kiros, and G.~E. Hinton, ``Layer normalization,'' \emph{arXiv},
  2016.

\bibitem{xiong2020layer}
R.~Xiong, Y.~Yang, D.~He, K.~Zheng, S.~Zheng, C.~Xing, H.~Zhang, Y.~Lan,
  L.~Wang, and T.~Liu, ``On layer normalization in the transformer
  architecture,'' in \emph{ICML}, 2020.

\bibitem{xu2021multigraph}
P.~Xu, C.~K. Joshi, and X.~Bresson, ``Multigraph transformer for free-hand
  sketch recognition,'' \emph{TNNLS}, 2021.

\bibitem{guo2021pct}
M.-H. Guo, J.-X. Cai, Z.-N. Liu, T.-J. Mu, R.~R. Martin, and S.-M. Hu, ``Pct:
  Point cloud transformer,'' \emph{Computational Visual Media}, 2021.

\bibitem{he2021transreid}
S.~He, H.~Luo, P.~Wang, F.~Wang, H.~Li, and W.~Jiang, ``Transreid:
  Transformer-based object re-identification,'' in \emph{ICCV}, 2021.

\bibitem{dufter2021position}
P.~Dufter, M.~Schmitt, and H.~Sch{\"u}tze, ``Position information in
  transformers: An overview,'' \emph{arXiv}, 2021.

\bibitem{wang2018non}
X.~Wang, R.~Girshick, A.~Gupta, and K.~He, ``Non-local neural networks,'' in
  \emph{CVPR}, 2018.

\bibitem{zhou2018end}
L.~Zhou, Y.~Zhou, J.~J. Corso, R.~Socher, and C.~Xiong, ``End-to-end dense
  video captioning with masked transformer,'' in \emph{CVPR}, 2018.

\bibitem{wang2019rat}
B.~Wang, R.~Shin, X.~Liu, O.~Polozov, and M.~Richardson, ``Rat-sql:
  Relation-aware schema encoding and linking for text-to-sql parsers,''
  \emph{arXiv}, 2019.

\bibitem{wang2021sgeitl}
Z.~Wang, H.~You, L.~H. Li, A.~Zareian, S.~Park, Y.~Liang, K.-W. Chang, and
  S.-F. Chang, ``Sgeitl: Scene graph enhanced image-text learning for visual
  commonsense reasoning,'' \emph{arXiv}, 2021.

\bibitem{glorot2011deep}
X.~Glorot, A.~Bordes, and Y.~Bengio, ``Deep sparse rectifier neural networks,''
  in \emph{AISTATS}, 2011.

\bibitem{hendrycks2016gaussian}
D.~Hendrycks and K.~Gimpel, ``Gaussian error linear units (gelus),''
  \emph{arXiv}, 2016.

\bibitem{lin2021exploring}
Y.-B. Lin, H.-Y. Tseng, H.-Y. Lee, Y.-Y. Lin, and M.-H. Yang, ``Exploring
  cross-video and cross-modality signals for weakly-supervised audio-visual
  video parsing,'' \emph{NeurIPS}, 2021.

\bibitem{akbari2021vatt}
H.~Akbari, L.~Yuan, R.~Qian, W.-H. Chuang, S.-F. Chang, Y.~Cui, and B.~Gong,
  ``Vatt: Transformers for multimodal self-supervised learning from raw video,
  audio and text,'' \emph{arXiv}, 2021.

\bibitem{nagrani2021attention}
A.~Nagrani, S.~Yang, A.~Arnab, A.~Jansen, C.~Schmid, and C.~Sun, ``Attention
  bottlenecks for multimodal fusion,'' \emph{NeurIPS}, 2021.

\bibitem{duquenne2021multimodal}
P.-A. Duquenne, H.~Gong, and H.~Schwenk, ``Multimodal and multilingual
  embeddings for large-scale speech mining,'' \emph{NeurIPS}, 2021.

\bibitem{min2021meta}
D.~Min, D.~B. Lee, E.~Yang, and S.~J. Hwang, ``Meta-stylespeech: Multi-speaker
  adaptive text-to-speech generation,'' \emph{arXiv}, 2021.

\bibitem{shi2022learning}
B.~Shi, W.-N. Hsu, K.~Lakhotia, and A.~Mohamed, ``Learning audio-visual speech
  representation by masked multimodal cluster prediction,'' \emph{arXiv}, 2022.

\bibitem{ramesh2021vset}
K.~Ramesh, C.~Xing, W.~Wang, D.~Wang, and X.~Chen, ``Vset: A multimodal
  transformer for visual speech enhancement,'' in \emph{ICASSP}, 2021.

\bibitem{zheng2021fused}
R.~Zheng, J.~Chen, M.~Ma, and L.~Huang, ``Fused acoustic and text encoding for
  multimodal bilingual pretraining and speech translation,'' \emph{arXiv},
  2021.

\bibitem{yang2021multimodal}
X.~Yang, S.~Feng, Y.~Zhang, and D.~Wang, ``Multimodal sentiment detection based
  on multi-channel graph neural networks,'' in \emph{ACL-IJCNLP}, 2021.

\bibitem{mao2021towards}
X.~Mao, G.~Qi, Y.~Chen, X.~Li, R.~Duan, S.~Ye, Y.~He, and H.~Xue, ``Towards
  robust vision transformer,'' \emph{arXiv}, 2021.

\bibitem{rahman2021tribert}
T.~Rahman, M.~Yang, and L.~Sigal, ``Tribert: Human-centric audio-visual
  representation learning,'' \emph{NeurIPS}, 2021.

\bibitem{rasmy2021med}
L.~Rasmy, Y.~Xiang, Z.~Xie, C.~Tao, and D.~Zhi, ``Med-bert: pretrained
  contextualized embeddings on large-scale structured electronic health records
  for disease prediction,'' \emph{NPJ digital medicine}, 2021.

\bibitem{chen2021multimodal}
R.~J. Chen, M.~Y. Lu, W.-H. Weng, T.~Y. Chen, D.~F. Williamson, T.~Manz,
  M.~Shady, and F.~Mahmood, ``Multimodal co-attention transformer for survival
  prediction in gigapixel whole slide images,'' in \emph{ICCV}, 2021.

\bibitem{xie2017rethinking}
S.~Xie, C.~Sun, J.~Huang, Z.~Tu, and K.~Murphy, ``Rethinking spatiotemporal
  feature learning for video understanding,'' \emph{arXiv}, 2017.

\bibitem{tran2018closer}
D.~Tran, H.~Wang, L.~Torresani, J.~Ray, Y.~LeCun, and M.~Paluri, ``A closer
  look at spatiotemporal convolutions for action recognition,'' in \emph{CVPR},
  2018.

\bibitem{lin2020interbert}
J.~Lin, A.~Yang, Y.~Zhang, J.~Liu, J.~Zhou, and H.~Yang, ``Interbert:
  Vision-and-language interaction for multi-modal pretraining,'' \emph{arXiv},
  2020.

\bibitem{tsai2019multimodal}
Y.-H.~H. Tsai, S.~Bai, P.~P. Liang, J.~Z. Kolter, L.-P. Morency, and
  R.~Salakhutdinov, ``Multimodal transformer for unaligned multimodal language
  sequences,'' in \emph{ACL}, 2019.

\bibitem{murahari2020large}
V.~Murahari, D.~Batra, D.~Parikh, and A.~Das, ``Large-scale pretraining for
  visual dialog: A simple state-of-the-art baseline,'' in \emph{ECCV}, 2020.

\bibitem{li2021scheduled}
Y.~Li, Y.~Pan, T.~Yao, J.~Chen, and T.~Mei, ``Scheduled sampling in
  vision-language pretraining with decoupled encoder-decoder network,'' in
  \emph{AAAI}, 2021.

\bibitem{kim2021vilt}
W.~Kim, B.~Son, and I.~Kim, ``Vilt: Vision-and-language transformer without
  convolution or region supervision,'' in \emph{ICML}, 2021.

\bibitem{yang2022vision}
J.~Yang, J.~Duan, S.~Tran, Y.~Xu, S.~Chanda, L.~Chen, B.~Zeng, T.~Chilimbi, and
  J.~Huang, ``Vision-language pre-training with triple contrastive learning,''
  \emph{arXiv}, 2022.

\bibitem{li2022grounded}
L.~H. Li, P.~Zhang, H.~Zhang, J.~Yang, C.~Li, Y.~Zhong, L.~Wang, L.~Yuan,
  L.~Zhang, J.-N. Hwang \emph{et~al.}, ``Grounded language-image
  pre-training,'' in \emph{CVPR}, 2022.

\bibitem{zhang2022glipv2}
H.~Zhang, P.~Zhang, X.~Hu, Y.-C. Chen, L.~Li, X.~Dai, L.~Wang, L.~Yuan, J.-N.
  Hwang, and J.~Gao, ``Glipv2: Unifying localization and vision-language
  understanding,'' \emph{NeurIPS}, 2022.

\bibitem{li2021semvlp}
C.~Li, M.~Yan, H.~Xu, F.~Luo, W.~Wang, B.~Bi, and S.~Huang, ``Semvlp:
  Vision-language pre-training by aligning semantics at multiple levels,''
  \emph{arXiv}, 2021.

\bibitem{miech2020end}
A.~Miech, J.-B. Alayrac, L.~Smaira, I.~Laptev, J.~Sivic, and A.~Zisserman,
  ``End-to-end learning of visual representations from uncurated instructional
  videos,'' in \emph{CVPR}, 2020.

\bibitem{han2022temporal}
T.~Han, W.~Xie, and A.~Zisserman, ``Temporal alignment networks for long-term
  video,'' in \emph{CVPR}, 2022.

\bibitem{wang2021simvlm}
Z.~Wang, J.~Yu, A.~W. Yu, Z.~Dai, Y.~Tsvetkov, and Y.~Cao, ``Simvlm: Simple
  visual language model pretraining with weak supervision,'' \emph{arXiv},
  2021.

\bibitem{chen2020mam}
J.~Chen, M.~Ma, R.~Zheng, and L.~Huang, ``Mam: Masked acoustic modeling for
  end-to-end speech-to-text translation,'' \emph{arXiv}, 2020.

\bibitem{golestani2021using}
M.~Golestani, S.~Z. Razavi, Z.~Borhanifard, F.~Tahmasebian, and H.~Faili,
  ``Using bert encoding and sentence-level language model for sentence
  ordering,'' in \emph{TSD}, 2021.

\bibitem{ren2015faster}
S.~Ren, K.~He, R.~Girshick, and J.~Sun, ``Faster r-cnn: Towards real-time
  object detection with region proposal networks,'' \emph{NeurIPS}, 2015.

\bibitem{huang2021seeing}
Z.~Huang, Z.~Zeng, Y.~Huang, B.~Liu, D.~Fu, and J.~Fu, ``Seeing out of the box:
  End-to-end pre-training for vision-language representation learning,'' in
  \emph{CVPR}, 2021.

\bibitem{liu2021kd}
Y.~Liu, C.~Wu, S.-y. Tseng, V.~Lal, X.~He, and N.~Duan, ``Kd-vlp: Improving
  end-to-end vision-and-language pretraining with object knowledge
  distillation,'' \emph{arXiv}, 2021.

\bibitem{ramesh2021zero}
A.~Ramesh, M.~Pavlov, G.~Goh, S.~Gray, C.~Voss, A.~Radford, M.~Chen, and
  I.~Sutskever, ``Zero-shot text-to-image generation,'' \emph{arXiv}, 2021.

\bibitem{zhou2021uc2}
M.~Zhou, L.~Zhou, S.~Wang, Y.~Cheng, L.~Li, Z.~Yu, and J.~Liu, ``Uc2: Universal
  cross-lingual cross-modal vision-and-language pre-training,'' in \emph{CVPR},
  2021.

\bibitem{hao2020towards}
W.~Hao, C.~Li, X.~Li, L.~Carin, and J.~Gao, ``Towards learning a generic agent
  for vision-and-language navigation via pre-training,'' in \emph{CVPR}, 2020.

\bibitem{xia2021xgpt}
Q.~Xia, H.~Huang, N.~Duan, D.~Zhang, L.~Ji, Z.~Sui, E.~Cui, T.~Bharti, and
  M.~Zhou, ``Xgpt: Cross-modal generative pre-training for image captioning,''
  in \emph{NLPCC}, 2021.

\bibitem{yao2021filip}
L.~Yao, R.~Huang, L.~Hou, G.~Lu, M.~Niu, H.~Xu, X.~Liang, Z.~Li, X.~Jiang, and
  C.~Xu, ``Filip: Fine-grained interactive language-image pre-training,''
  \emph{arXiv}, 2021.

\bibitem{gan2020large}
Z.~Gan, Y.-C. Chen, L.~Li, C.~Zhu, Y.~Cheng, and J.~Liu, ``Large-scale
  adversarial training for vision-and-language representation learning,''
  \emph{arXiv}, 2020.

\bibitem{zhang2021ernie}
H.~Zhang, W.~Yin, Y.~Fang, L.~Li, B.~Duan, Z.~Wu, Y.~Sun, H.~Tian, H.~Wu, and
  H.~Wang, ``Ernie-vilg: Unified generative pre-training for bidirectional
  vision-language generation,'' \emph{arXiv}, 2021.

\bibitem{zhuge2021kaleido}
M.~Zhuge, D.~Gao, D.-P. Fan, L.~Jin, B.~Chen, H.~Zhou, M.~Qiu, and L.~Shao,
  ``Kaleido-bert: Vision-language pre-training on fashion domain,'' in
  \emph{CVPR}, 2021.

\bibitem{wang2022multimodal}
Y.~Wang, X.~Chen, L.~Cao, W.~Huang, F.~Sun, and Y.~Wang, ``Multimodal token
  fusion for vision transformers,'' in \emph{CVPR}, 2022.

\bibitem{wang2022bridged}
Y.~Wang, T.~Ye, L.~Cao, W.~Huang, F.~Sun, F.~He, and D.~Tao, ``Bridged
  transformer for vision and point cloud 3d object detection,'' in \emph{CVPR},
  2022.

\bibitem{prakash2021multi}
A.~Prakash, K.~Chitta, and A.~Geiger, ``Multi-modal fusion transformer for
  end-to-end autonomous driving,'' in \emph{CVPR}, 2021.

\bibitem{bai2022transfusion}
X.~Bai, Z.~Hu, X.~Zhu, Q.~Huang, Y.~Chen, H.~Fu, and C.-L. Tai, ``Transfusion:
  Robust lidar-camera fusion for 3d object detection with transformers,'' in
  \emph{CVPR}, 2022.

\bibitem{favory2021learning}
X.~Favory, K.~Drossos, T.~Virtanen, and X.~Serra, ``Learning contextual tag
  embeddings for cross-modal alignment of audio and tags,'' in \emph{ICASSP},
  2021.

\bibitem{gabeur2020multi}
V.~Gabeur, C.~Sun, K.~Alahari, and C.~Schmid, ``Multi-modal transformer for
  video retrieval,'' in \emph{ECCV}, 2020.

\bibitem{Shvetsova_2022_CVPRMulti}
N.~Shvetsova, B.~Chen, A.~Rouditchenko, S.~Thomas, B.~Kingsbury, R.~S. Feris,
  D.~Harwath, J.~Glass, and H.~Kuehne, ``Everything at once - multi-modal
  fusion transformer for video retrieval,'' in \emph{CVPR}, 2022.

\bibitem{wang2022multi}
Z.~Wang, Y.~Wu, K.~Narasimhan, and O.~Russakovsky, ``Multi-query video
  retrieval,'' \emph{arXiv}, 2022.

\bibitem{botach2021end}
A.~Botach, E.~Zheltonozhskii, and C.~Baskin, ``End-to-end referring video
  object segmentation with multimodal transformers,'' \emph{arXiv}, 2021.

\bibitem{hong2021gilbert}
W.~Hong, K.~Ji, J.~Liu, J.~Wang, J.~Chen, and W.~Chu, ``Gilbert: Generative
  vision-language pre-training for image-text retrieval,'' in \emph{SIGIR},
  2021.

\bibitem{xu20213d}
X.~Xu and C.~C. Loy, ``3d human texture estimation from a single image with
  transformers,'' in \emph{ICCV}, 2021.

\bibitem{lin2020gps}
X.~Lin, C.~Ding, J.~Zeng, and D.~Tao, ``Gps-net: Graph property sensing network
  for scene graph generation,'' in \emph{CVPR}, 2020.

\bibitem{wang2021topic}
W.~Wang, R.~Wang, and X.~Chen, ``Topic scene graph generation by attention
  distillation from caption,'' in \emph{ICCV}, 2021.

\bibitem{lu2021context}
Y.~Lu, H.~Rai, J.~Chang, B.~Knyazev, G.~Yu, S.~Shekhar, G.~W. Taylor, and
  M.~Volkovs, ``Context-aware scene graph generation with seq2seq
  transformers,'' in \emph{ICCV}, 2021.

\bibitem{ke2021jointgt}
P.~Ke, H.~Ji, Y.~Ran, X.~Cui, L.~Wang, L.~Song, X.~Zhu, and M.~Huang,
  ``Jointgt: Graph-text joint representation learning for text generation from
  knowledge graphs,'' \emph{arXiv}, 2021.

\bibitem{teng2021target}
Y.~Teng, L.~Wang, Z.~Li, and G.~Wu, ``Target adaptive context aggregation for
  video scene graph generation,'' in \emph{ICCV}, 2021.

\bibitem{chen2018tvt}
M.~Chen, Y.~Li, Z.~Zhang, and S.~Huang, ``Tvt: Two-view transformer network for
  video captioning,'' in \emph{ACML}, 2018.

\bibitem{lin2021swinbert}
K.~Lin, L.~Li, C.-C. Lin, F.~Ahmed, Z.~Gan, Z.~Liu, Y.~Lu, and L.~Wang,
  ``Swinbert: End-to-end transformers with sparse attention for video
  captioning,'' \emph{arXiv}, 2021.

\bibitem{deng2021sketch}
C.~Deng, S.~Chen, D.~Chen, Y.~He, and Q.~Wu, ``Sketch, ground, and refine:
  Top-down dense video captioning,'' in \emph{CVPR}, 2021.

\bibitem{wang2021end}
T.~Wang, R.~Zhang, Z.~Lu, F.~Zheng, R.~Cheng, and P.~Luo, ``End-to-end dense
  video captioning with parallel decoding,'' in \emph{ICCV}, 2021.

\bibitem{huang2019attention}
L.~Huang, W.~Wang, J.~Chen, and X.-Y. Wei, ``Attention on attention for image
  captioning,'' in \emph{ICCV}, 2019.

\bibitem{pan2020x}
Y.~Pan, T.~Yao, Y.~Li, and T.~Mei, ``X-linear attention networks for image
  captioning,'' in \emph{CVPR}, 2020.

\bibitem{yang2021causal}
X.~Yang, H.~Zhang, G.~Qi, and J.~Cai, ``Causal attention for vision-language
  tasks,'' in \emph{CVPR}, 2021.

\bibitem{luo2021dual}
Y.~Luo, J.~Ji, X.~Sun, L.~Cao, Y.~Wu, F.~Huang, C.-W. Lin, and R.~Ji,
  ``Dual-level collaborative transformer for image captioning,'' \emph{arXiv},
  2021.

\bibitem{xu2021towards}
G.~Xu, S.~Niu, M.~Tan, Y.~Luo, Q.~Du, and Q.~Wu, ``Towards accurate text-based
  image captioning with content diversity exploration,'' in \emph{CVPR}, 2021.

\bibitem{li2019neural}
N.~Li, S.~Liu, Y.~Liu, S.~Zhao, and M.~Liu, ``Neural speech synthesis with
  transformer network,'' in \emph{AAAI}, 2019.

\bibitem{ding2021cogview}
M.~Ding, Z.~Yang, W.~Hong, W.~Zheng, C.~Zhou, D.~Yin, J.~Lin, X.~Zou, Z.~Shao,
  H.~Yang \emph{et~al.}, ``Cogview: Mastering text-to-image generation via
  transformers,'' \emph{arXiv}, 2021.

\bibitem{sanghi2021clip}
A.~Sanghi, H.~Chu, J.~G. Lambourne, Y.~Wang, C.-Y. Cheng, and M.~Fumero,
  ``Clip-forge: Towards zero-shot text-to-shape generation,'' \emph{arXiv},
  2021.

\bibitem{huang2020dance}
R.~Huang, H.~Hu, W.~Wu, K.~Sawada, M.~Zhang, and D.~Jiang, ``Dance revolution:
  Long-term dance generation with music via curriculum learning,''
  \emph{arXiv}, 2020.

\bibitem{zhong2021learning}
Y.~Zhong, J.~Shi, J.~Yang, C.~Xu, and Y.~Li, ``Learning to generate scene graph
  from natural language supervision,'' in \emph{ICCV}, 2021.

\bibitem{geng2021dynamic}
S.~Geng, P.~Gao, M.~Chatterjee, C.~Hori, J.~Le~Roux, Y.~Zhang, H.~Li, and
  A.~Cherian, ``Dynamic graph representation learning for video dialog via
  multi-modal shuffled transformers,'' in \emph{AAAI}, 2021.

\bibitem{jaunet2021visqa}
T.~Jaunet, C.~Kervadec, R.~Vuillemot, G.~Antipov, M.~Baccouche, and C.~Wolf,
  ``Visqa: X-raying vision and language reasoning in transformers,''
  \emph{TVCG}, 2021.

\bibitem{lin2021vx2text}
X.~Lin, G.~Bertasius, J.~Wang, S.-F. Chang, D.~Parikh, and L.~Torresani,
  ``Vx2text: End-to-end learning of video-based text generation from multimodal
  inputs,'' in \emph{CVPR}, 2021.

\bibitem{lin2021m6}
J.~Lin, R.~Men, A.~Yang, C.~Zhou, Y.~Zhang, P.~Wang, J.~Zhou, J.~Tang, and
  H.~Yang, ``M6: Multi-modality-to-multi-modality multitask mega-transformer
  for unified pretraining,'' in \emph{KDD}, 2021.

\bibitem{chen1998audio}
T.~Chen and R.~R. Rao, ``Audio-visual integration in multimodal
  communication,'' \emph{Proceedings of the IEEE}, 1998.

\bibitem{owens2018audio}
A.~Owens and A.~A. Efros, ``Audio-visual scene analysis with self-supervised
  multisensory features,'' in \emph{ECCV}, 2018.

\bibitem{xue2021probing}
H.~Xue, Y.~Huang, B.~Liu, H.~Peng, J.~Fu, H.~Li, and J.~Luo, ``Probing
  inter-modality: Visual parsing with self-attention for vision-and-language
  pre-training,'' \emph{NeurIPS}, 2021.

\bibitem{truong2021right}
T.-D. Truong, C.~N. Duong, H.~A. Pham, B.~Raj, N.~Le, K.~Luu \emph{et~al.},
  ``The right to talk: An audio-visual transformer approach,'' in \emph{ICCV},
  2021.

\bibitem{chen2020multispeech}
M.~Chen, X.~Tan, Y.~Ren, J.~Xu, H.~Sun, S.~Zhao, T.~Qin, and T.-Y. Liu,
  ``Multispeech: Multi-speaker text to speech with transformer,'' \emph{arXiv},
  2020.

\bibitem{ging2020coot}
S.~Ging, M.~Zolfaghari, H.~Pirsiavash, and T.~Brox, ``Coot: Cooperative
  hierarchical transformer for video-text representation learning,''
  \emph{arXiv}, 2020.

\bibitem{patrick2020support}
M.~Patrick, P.-Y. Huang, Y.~Asano, F.~Metze, A.~Hauptmann, J.~Henriques, and
  A.~Vedaldi, ``Support-set bottlenecks for video-text representation
  learning,'' \emph{arXiv}, 2020.

\bibitem{gabeur2022masking}
V.~Gabeur, A.~Nagrani, C.~Sun, K.~Alahari, and C.~Schmid, ``Masking modalities
  for cross-modal video retrieval,'' in \emph{WACV}, 2022.

\bibitem{sadhu2020video}
A.~Sadhu, K.~Chen, and R.~Nevatia, ``Video object grounding using semantic
  roles in language description,'' in \emph{CVPR}, 2020.

\bibitem{zhang2021explainable}
Y.~Zhang, M.~Choi, K.~Han, and Z.~Liu, ``Explainable semantic space by
  grounding language to vision with cross-modal contrastive learning,''
  \emph{NeurIPS}, 2021.

\bibitem{chen2021end}
Y.-W. Chen, Y.-H. Tsai, and M.-H. Yang, ``End-to-end multi-modal video temporal
  grounding,'' \emph{NeurIPS}, 2021.

\bibitem{Chen_2022_CVPR}
S.~Chen and B.~Li, ``Multi-modal dynamic graph transformer for visual
  grounding,'' in \emph{CVPR}, 2022.

\bibitem{Yang_2022_CVPR_TubeDETR}
A.~Yang, A.~Miech, J.~Sivic, I.~Laptev, and C.~Schmid, ``Tubedetr:
  Spatio-temporal video grounding with transformers,'' in \emph{CVPR}, 2022.

\bibitem{xu2021videoclip}
H.~Xu, G.~Ghosh, P.-Y. Huang, D.~Okhonko, A.~Aghajanyan, F.~Metze,
  L.~Zettlemoyer, and C.~Feichtenhofer, ``Videoclip: Contrastive pre-training
  for zero-shot video-text understanding,'' \emph{arXiv}, 2021.

\bibitem{lei2021less}
J.~Lei, L.~Li, L.~Zhou, Z.~Gan, T.~L. Berg, M.~Bansal, and J.~Liu, ``Less is
  more: Clipbert for video-and-language learning via sparse sampling,'' in
  \emph{CVPR}, 2021.

\bibitem{yang2021taco}
J.~Yang, Y.~Bisk, and J.~Gao, ``Taco: Token-aware cascade contrastive learning
  for video-text alignment,'' in \emph{ICCV}, 2021.

\bibitem{li2021align}
D.~Li, J.~Li, H.~Li, J.~C. Niebles, and S.~C. Hoi, ``Align and prompt:
  Video-and-language pre-training with entity prompts,'' \emph{arXiv}, 2021.

\bibitem{luo2021clip4clip}
H.~Luo, L.~Ji, M.~Zhong, Y.~Chen, W.~Lei, N.~Duan, and T.~Li, ``Clip4clip: An
  empirical study of clip for end to end video clip retrieval,'' \emph{arXiv},
  2021.

\bibitem{fang2021clip2video}
H.~Fang, P.~Xiong, L.~Xu, and Y.~Chen, ``Clip2video: Mastering video-text
  retrieval via image clip,'' \emph{arXiv}, 2021.

\bibitem{narasimhan2021clip}
M.~Narasimhan, A.~Rohrbach, and T.~Darrell, ``Clip-it! language-guided video
  summarization,'' \emph{NeurIPS}, 2021.

\bibitem{liu2023pre}
P.~Liu, W.~Yuan, J.~Fu, Z.~Jiang, H.~Hayashi, and G.~Neubig, ``Pre-train,
  prompt, and predict: A systematic survey of prompting methods in natural
  language processing,'' \emph{ACM Computing Surveys}, 2023.

\bibitem{xian2018zero}
Y.~Xian, C.~H. Lampert, B.~Schiele, and Z.~Akata, ``Zero-shot learning—a
  comprehensive evaluation of the good, the bad and the ugly,'' \emph{TPAMI},
  2018.

\bibitem{gu2021open}
X.~Gu, T.-Y. Lin, W.~Kuo, and Y.~Cui, ``Open-vocabulary object detection via
  vision and language knowledge distillation,'' \emph{arXiv}, 2021.

\bibitem{cho2020x}
J.~Cho, J.~Lu, D.~Schwenk, H.~Hajishirzi, and A.~Kembhavi, ``X-lxmert: Paint,
  caption and answer questions with multi-modal transformers,'' in
  \emph{EMNLP}, 2020.

\bibitem{xu2021e2e}
H.~Xu, M.~Yan, C.~Li, B.~Bi, S.~Huang, W.~Xiao, and F.~Huang, ``E2e-vlp:
  End-to-end vision-language pre-training enhanced by visual learning,''
  \emph{arXiv}, 2021.

\bibitem{kervadec2021supervising}
C.~Kervadec, C.~Wolf, G.~Antipov, M.~Baccouche, and M.~Nadri, ``Supervising the
  transfer of reasoning patterns in vqa,'' \emph{arXiv}, 2021.

\bibitem{kervadec2021transferable}
C.~Kervadec, T.~Jaunet, G.~Antipov, M.~Baccouche, R.~Vuillemot, and C.~Wolf,
  ``How transferable are reasoning patterns in vqa?'' in \emph{CVPR}, 2021.

\bibitem{rahman2020integrating}
W.~Rahman, M.~K. Hasan, S.~Lee, A.~Zadeh, C.~Mao, L.-P. Morency, and E.~Hoque,
  ``Integrating multimodal information in large pretrained transformers,'' in
  \emph{ACL}, 2020.

\bibitem{agarwal2021multimodal}
D.~Agarwal, T.~Agrawal, L.~M. Ferrari, and F.~Bremond, ``From multimodal to
  unimodal attention in transformers using knowledge distillation,'' in
  \emph{AVSS}, 2021.

\bibitem{li2021towards}
Q.~Li, B.~Gong, Y.~Cui, D.~Kondratyuk, X.~Du, M.-H. Yang, and M.~Brown,
  ``Towards a unified foundation model: Jointly pre-training transformers on
  unpaired images and text,'' \emph{arXiv}, 2021.

\bibitem{ni2021m3p}
M.~Ni, H.~Huang, L.~Su, E.~Cui, T.~Bharti, L.~Wang, D.~Zhang, and N.~Duan,
  ``M3p: Learning universal representations via multitask multilingual
  multimodal pre-training,'' in \emph{CVPR}, 2021.

\bibitem{miech2021thinking}
A.~Miech, J.-B. Alayrac, I.~Laptev, J.~Sivic, and A.~Zisserman, ``Thinking fast
  and slow: Efficient text-to-visual retrieval with transformers,'' in
  \emph{CVPR}, 2021.

\bibitem{wen2021cookie}
K.~Wen, J.~Xia, Y.~Huang, L.~Li, J.~Xu, and J.~Shao, ``Cookie: Contrastive
  cross-modal knowledge sharing pre-training for vision-language
  representation,'' in \emph{ICCV}, 2021.

\bibitem{lee2020parameter}
S.~Lee, Y.~Yu, G.~Kim, T.~Breuel, J.~Kautz, and Y.~Song, ``Parameter efficient
  multimodal transformers for video representation learning,'' \emph{arXiv},
  2020.

\bibitem{liu2021multi}
T.~Liu, F.~Feng, and X.~Wang, ``Multi-stage pre-training over simplified
  multimodal pre-training models,'' \emph{arXiv}, 2021.

\bibitem{li2021supervision}
Y.~Li, F.~Liang, L.~Zhao, Y.~Cui, W.~Ouyang, J.~Shao, F.~Yu, and J.~Yan,
  ``Supervision exists everywhere: A data efficient contrastive language-image
  pre-training paradigm,'' \emph{arXiv}, 2021.

\bibitem{gan2021playing}
Z.~Gan, Y.-C. Chen, L.~Li, T.~Chen, Y.~Cheng, S.~Wang, and J.~Liu, ``Playing
  lottery tickets with vision and language,'' \emph{arXiv}, 2021.

\bibitem{he2021pipetransformer}
C.~He, S.~Li, M.~Soltanolkotabi, and S.~Avestimehr, ``Pipetransformer:
  Automated elastic pipelining for distributed training of large-scale
  models,'' in \emph{ICML}, 2021.

\bibitem{dao2022flashattention}
T.~Dao, D.~Fu, S.~Ermon, A.~Rudra, and C.~R{\'e}, ``Flashattention: Fast and
  memory-efficient exact attention with io-awareness,'' \emph{NeurIPS}, 2022.

\bibitem{child2019generating}
R.~Child, S.~Gray, A.~Radford, and I.~Sutskever, ``Generating long sequences
  with sparse transformers,'' \emph{arXiv}, 2019.

\bibitem{zhu2021long}
C.~Zhu, W.~Ping, C.~Xiao, M.~Shoeybi, T.~Goldstein, A.~Anandkumar, and
  B.~Catanzaro, ``Long-short transformer: Efficient transformers for language
  and vision,'' in \emph{NeurIPS}, 2021.

\bibitem{yan2022multiview}
S.~Yan, X.~Xiong, A.~Arnab, Z.~Lu, M.~Zhang, C.~Sun, and C.~Schmid, ``Multiview
  transformers for video recognition,'' in \emph{CVPR}, 2022.

\bibitem{wang2020rethinking}
W.~Wang and Z.~Tu, ``Rethinking the value of transformer components,'' in
  \emph{COLING}, 2020.

\bibitem{Ma_2022_CVPR_are_multimodal}
M.~Ma, J.~Ren, L.~Zhao, D.~Testuggine, and X.~Peng, ``Are multimodal
  transformers robust to missing modality?'' in \emph{CVPR}, 2022.

\bibitem{akula2021robust}
A.~Akula, V.~Jampani, S.~Changpinyo, and S.-C. Zhu, ``Robust visual reasoning
  via language guided neural module networks,'' \emph{NeurIPS}, 2021.

\bibitem{akula2020words}
A.~R. Akula, S.~Gella, Y.~Al-Onaizan, S.-C. Zhu, and S.~Reddy, ``Words aren't
  enough, their order matters: On the robustness of grounding visual referring
  expressions,'' \emph{arXiv}, 2020.

\bibitem{li2020closer}
L.~Li, Z.~Gan, and J.~Liu, ``A closer look at the robustness of
  vision-and-language pre-trained models,'' \emph{arXiv}, 2020.

\bibitem{zhang2021domain}
M.~Zhang, T.~Maidment, A.~Diab, A.~Kovashka, and R.~Hwa, ``Domain-robust vqa
  with diverse datasets and methods but no target labels,'' in \emph{CVPR},
  2021.

\bibitem{kant2021contrast}
Y.~Kant, A.~Moudgil, D.~Batra, D.~Parikh, and H.~Agrawal, ``Contrast and
  classify: Training robust vqa models,'' in \emph{ICCV}, 2021.

\bibitem{pramanik2019omninet}
S.~Pramanik, P.~Agrawal, and A.~Hussain, ``Omninet: A unified architecture for
  multi-modal multi-task learning,'' \emph{arXiv}, 2019.

\bibitem{wang2022unifying}
P.~Wang, A.~Yang, R.~Men, J.~Lin, S.~Bai, Z.~Li, J.~Ma, C.~Zhou, J.~Zhou, and
  H.~Yang, ``Unifying architectures, tasks, and modalities through a simple
  sequence-to-sequence learning framework,'' \emph{arXiv}, 2022.

\bibitem{girdhar2022omnivore}
R.~Girdhar, M.~Singh, N.~Ravi, L.~van~der Maaten, A.~Joulin, and I.~Misra,
  ``Omnivore: A single model for many visual modalities,'' \emph{arXiv}, 2022.

\bibitem{cao2020behind}
J.~Cao, Z.~Gan, Y.~Cheng, L.~Yu, Y.-C. Chen, and J.~Liu, ``Behind the scene:
  Revealing the secrets of pre-trained vision-and-language models,'' in
  \emph{ECCV}, 2020.

\bibitem{hendricks2021decoupling}
L.~A. Hendricks, J.~Mellor, R.~Schneider, J.-B. Alayrac, and A.~Nematzadeh,
  ``Decoupling the role of data, attention, and losses in multimodal
  transformers,'' \emph{TACL}, 2021.

\bibitem{hendricks2021probing}
L.~A. Hendricks and A.~Nematzadeh, ``Probing image-language transformers for
  verb understanding,'' \emph{arXiv}, 2021.

\bibitem{frank2021vision}
S.~Frank, E.~Bugliarello, and D.~Elliott, ``Vision-and-language or
  vision-for-language? on cross-modal influence in multimodal transformers,''
  \emph{arXiv}, 2021.

\bibitem{chefer2021generic}
H.~Chefer, S.~Gur, and L.~Wolf, ``Generic attention-model explainability for
  interpreting bi-modal and encoder-decoder transformers,'' \emph{arXiv}, 2021.

\bibitem{parcalabescu2021valse}
L.~Parcalabescu, M.~Cafagna, L.~Muradjan, A.~Frank, I.~Calixto, and A.~Gatt,
  ``Valse: A task-independent benchmark for vision and language models centered
  on linguistic phenomena,'' \emph{arXiv}, 2021.

\bibitem{zhao2022vl}
T.~Zhao, T.~Zhang, M.~Zhu, H.~Shen, K.~Lee, X.~Lu, and J.~Yin, ``Vl-checklist:
  Evaluating pre-trained vision-language models with objects, attributes and
  relations,'' \emph{arXiv}, 2022.

\bibitem{Aflalo_2022_CVPR}
E.~Aflalo, M.~Du, S.-Y. Tseng, Y.~Liu, C.~Wu, N.~Duan, and V.~Lal,
  ``Vl-interpret: An interactive visualization tool for interpreting
  vision-language transformers,'' in \emph{CVPR}, 2022.

\bibitem{mu2021slip}
N.~Mu, A.~Kirillov, D.~Wagner, and S.~Xie, ``Slip: Self-supervision meets
  language-image pre-training,'' \emph{arXiv}, 2021.

\bibitem{xu2021vlm}
H.~Xu, G.~Ghosh, P.-Y. Huang, P.~Arora, M.~Aminzadeh, C.~Feichtenhofer,
  F.~Metze, and L.~Zettlemoyer, ``Vlm: Task-agnostic video-language model
  pre-training for video understanding,'' \emph{arXiv}, 2021.

\bibitem{li2022blip}
J.~Li, D.~Li, C.~Xiong, and S.~Hoi, ``Blip: Bootstrapping language-image
  pre-training for unified vision-language understanding and generation,''
  \emph{arXiv}, 2022.

\bibitem{zhang2021vinvl}
P.~Zhang, X.~Li, X.~Hu, J.~Yang, L.~Zhang, L.~Wang, Y.~Choi, and J.~Gao,
  ``Vinvl: Revisiting visual representations in vision-language models,'' in
  \emph{CVPR}, 2021.

\end{thebibliography}
